\documentclass[a4paper,fleqn]{cas-dc}

\usepackage[utf8]{inputenc}
\usepackage{graphicx}
\usepackage{float}

\usepackage[compress]{natbib}
\usepackage{pifont}

\usepackage{caption} 
\usepackage{subcaption} 

\usepackage{cleveref}

\usepackage{lipsum}
\usepackage{multicol}

\usepackage{times}
\usepackage{epsfig}
\usepackage{amsmath}
\usepackage{amssymb}
\usepackage{booktabs}
\usepackage{multirow}
\usepackage{pifont}

\usepackage{svg}
\usepackage{xcolor}
\usepackage{amsthm}
\usepackage{dirtytalk}
\usepackage{hyperref}
\usepackage{url}

\usepackage{amsmath,amsfonts,bm}

\def\eqref#1{equation~\ref{#1}}

\def\1{\bm{1}}

\def\vx{{\bm{x}}}
\def\vy{{\bm{y}}}
\def\vz{{\bm{z}}}

\DeclareMathAlphabet{\mathsfit}{\encodingdefault}{\sfdefault}{m}{sl}
\SetMathAlphabet{\mathsfit}{bold}{\encodingdefault}{\sfdefault}{bx}{n}

\def\tsc#1{\csdef{#1}{\textsc{\lowercase{#1}}\xspace}}
\tsc{WGM}
\tsc{QE}
\tsc{EP}
\tsc{PMS}
\tsc{BEC}
\tsc{DE}

\begin{document}
\let\WriteBookmarks\relax
\def\floatpagepagefraction{1}
\def\textpagefraction{.001}

\shorttitle{A metric learning approach for endoscopic kidney stone identification}

\shortauthors{J. Gonzalez-Zapata et~al.}

\title [mode = title]{A metric learning approach for endoscopic kidney stone identification}   

\author[addr1]{Jorge Gonzalez-Zapata}
\author[addr2,addr3]{Francisco Lopez-Tiro}
\author[addr2]{Elias Villalvazo-Avila}
\author[addr2]{Daniel Flores-Araiza}
\author[addr4]{Jacques Hubert}
\author[addr1]{Andres Mendez-Vazquez}
\author[addr2]{Gilberto Ochoa-Ruiz*}
\author[addr3]{Christian Daul*}

\cortext[cor]{Corresponding author: andres.mendez@cinvestav.mx, \\ gilberto.ochoa@tec.mx, christian.daul@univ-lorraine.fr}

\address[addr1]{Centro de Investigación y de Estudios Avanzados, Computer Sciences Department, Guadalajara, Mexico}
\address[addr2]{Tecnologico de Monterrey, School of Engineering and Sciences, Mexico}
\address[addr3]{Centre de Recherche en Automatique de Nancy (UMR 7039, CNRS and Université de Lorraine),  Vand{\oe}uvre-l\`es-Nancy, France}
\address[addr4]{CHU Nancy, Service d’urologie de Brabois, Vand{\oe}uvre-l\`es-Nancy, France}

\maketitle

\begin{abstract}
Several Deep Learning (DL) methods have recently been proposed for an automated identification of kidney stones during an ureteroscopy to enable rapid therapeutic decisions. Even if these DL approaches led to promising results, they are mainly appropriate for kidney stone types for which numerous labelled data are available. However, only few labelled images are available for some rare kidney stone types. This contribution exploits Deep Metric Learning (DML) methods i) to handle such classes with few samples, ii) to generalize well to out of distribution samples, and iii) to cope better with new classes which are added to the database. The proposed \textit{Guided Deep Metric Learning} approach is based on a novel architecture which was designed to learn data representations in an improved way. The solution was inspired by Few-Shot Learning (FSL) and makes use of a teacher-student approach. The teacher model (GEMINI) generates a reduced hypothesis space based on prior knowledge from the labeled data, and is used it as a \say{guide} to a student model (i.e., ResNet50) through a Knowledge Distillation scheme.
Extensive tests were first performed on two datasets separately used for the recognition, namely a set of images acquired for the surfaces of the kidney stone fragments, and a set of images of the fragment sections.  
The proposed DML-approach improved the identification accuracy by 10\% and 12\% in comparison to DL-methods and other DML-approaches, respectively. Moreover, model embeddings from the two dataset types were merged in an organized way through a multi-view scheme to simultaneously exploit the information of surface and section fragments. Test with the resulting mixed model improves the identification accuracy by at least 3\% and up to 30\% with respect to DL-models and shallow machine learning methods, respectively.

\end{abstract}



\section{Introduction}
\label{sec:introduction}


\subsection{Medical context} \label{sec:MedicalContext}

Renal lithiasis is the ultimate stage in the kidney stone formation by the concretion of crystalline particles. 
Small-sized kidney stones are able to naturally leave the urinary tract (\cite{cloutier2015}). However, larger kidney stones (i.e., with a dia\-meter exceeding  some millimeters) cannot naturally drain and cause severe pain. Such larger kidney stones are often removed from the urinary tract during an ureteroscopy using a flexible endoscope.
In industrialized countries, there is a high incidence of renal lithiasis since up to 10\% of the population suffers from such an event at least once in their lifetime (\cite{kasidas2004, hall2009}). The recurrence rate of urinary calculi can reach up to 40\% without an appropriate anti-recurrence  treatment (\cite{scales2012, viljoen2019}).
Diet is among the most common factors in the formation of kidney stones (\cite{friedlander2015}). 
Furthermore, there is a relationship between recurrence risk factors and the biochemical composition of kidney stones (\cite{silva2010, daudon2012}).
Thus, a reliable and automated identification of the type of kidney stones is crucial in order to determine a personalized treatment and to prevent possible relapses (\cite{kartha2013, friedlander2015}).

The current reference method to categorize a kidney stone type is the ``Morpho-Constitutional Analysis'' (MCA, (\cite{daudon2012})). During an ureteroscopy, kidney stones are fragmented using a laser and extracted from the urinary tract. The surface and section of the kidney stone fragments are visually analyzed by biologists in terms of colors, textures, and crystalline morphology. The fragments are then reduced to powder and a FTIR (Fourier Transform Infrared Spectroscopy) analysis provides information about the biochemical composition of the fragments. MCA is accurate but, for operational reasons of hospitals, the results of this ex-vivo analysis are only available some weeks or months after the ureteroscopy. Since for some calculi types an immediate treatment is recommended, some urologists have trained themselves to visually recognize the kidney stone types on the video displayed on a screen during the ureteroscopy. However, this in-vivo technique (called Endoscopic Stone Recognition, ESR, (\cite{estrade2013}) requires a great deal of experience and is operator dependent (\cite{sampogna2020}).         

\subsection{Automated kidney-stone type recognition \label{sec:PreviousWorks}} 
Given the importance of a fast anti-recurrence  treatment, several methods to automate the kidney stone identification have been proposed in the literature. Some initial attempts made use of Shallow Machine Learning (SML) methods and subsequently, Deep Learning (DL) based approaches have been investigated. These works are discussed in the following sections, highlighting their strengths and limitations.
\subsubsection{Shallow machine learning in ureteroscopy \label{sec:PreviousWorksSML}} 
Similarly to the visual description performed by biologists during MCA, the first two works on automated kidney stone recognition exploited information relating to the texture, color and morphology of fragments. These handcrafted features (based on \say{mathematically} interpretable image content) were treated by various classifiers. 

In the first contribution dealing with kidney stone recognition (\cite{serrat2017}), color and texture information were respectively stored in histograms of  RGB-values and local binary patterns invariant to rotations. These feature vectors were used to train a random forest classifier. The data were obtained in ex-vivo since digital cameras were used to acquire images of the extracted fragments in a environment ensuring optimal scene illumination and controlled viewpoints, as well as contrasted and blurless image content. 
The work in (\cite{martinez2020}) used in-vivo data (kidney stones after fragmentation) acquired with an endoscope during an ureteroscopy. For such in-situ data the ureteroscope's viewpoint is difficult to control, the illumination conditions present strong variations, and the images can be blurred. One contribution of this work (based on an ensemble of KNN-classifiers) was to show that the HSI (Hue/Saturation/Intensity, (\cite{daul2000})) color space leads to more discriminant features than the RGB color space. The choice of appropriate color features allowed to pass from a 63\% classification accuracy in (\cite{serrat2017}) to a 88,5\% accuracy in (\cite{martinez2020}). 

The encouraging results of these two first contributions have shown the feasibility of an automated kidney stone recognition. It is worth noticing that section and surface fragment data were used separately by both contribution to identify the kidney stone types.    


\subsubsection{Deep learning in ureteroscopy \label{sec:PreviousWorksDL}} 


The first DL-approach for kidney stone identification was based on DML, (\cite{torrell2018metric})). The selected architecture (a Siamese network exploiting a ResNet-50) fused the image data of surface and section views. Although the idea of an architecture supporting fusion methods for both views was novel, the results (classification accuracy of 74\%) for ex-vivo data were less promising than that of the SML method of (\cite{martinez2020}) on in-vivo data. 
%

Two years later, (\cite{black2020}) and (\cite{estrade2020}) proposed other deep neural networks still based on a ResNet architecture (ResNet-101 for (\cite{black2020}) and ResNet-152-V2 for (\cite{estrade2020})). The results obtained for individual views were promising since (\cite{black2020}) classified ex-vivo images of five kidney stone sub-types with an accuracy from 71,43\% up to 95\% according to the class, and the authors in (\cite{estrade2020}) were able to distinguish between four classes of in-vivo images. 

In contrast to these previous works, in (\cite{estrade2020}) two classes of kidney stones contained two different biochemical components, while the other two types were pure (i.e. only with one biochemical component). The two pure sub-types were classified with an high accuracy of 91\% and 98\%, respectively. Neither (\cite{black2020}), nor  (\cite{estrade2020}) fused the fragment and surface data to improve the recognition. 

Some authors ((\cite{lopez2021assessing}) and (\cite{ochoa2022vivo})) proposed not only to use section and surface images separately in dedicated DL-architectures, but also trained DL-models by jointly using both views. These authors tested the joint use of both image types on architectures with different depths, namely AlexNet, ResNet-50, Inception-V3, and VGG16. The results obtained showed that deep neural networks combining both views extract more discriminant features. The feature space in mixed models show more compact class clusters compared to those of the individual surface and section views. Thus (\cite{lopez2021assessing}) report a precision and recall of 97\% and 98\%, respectively, obtained over four classes (i.e., for in-vivo images of four kidney stone sub-types). 

In more recent works, Metric Learning approaches were used to take into account the rather small size of the kidney stone image databases available to train the DL-networks. For instance, Few-Shot Learning (FSL) and Meta-Learning techniques were used in (\cite{mendez2022generalization}) to classify kidney stone fragments. In this work, a Meta Learning scheme that refines weights of large ensembles (such as ImageNet) under an FSL scheme is employed. This approach based on Meta Learning and FSL avoids training directly with kidney stone images.  For evaluating the model, the weights obtained through the meta learning optimization process are used. The results obtained on images for individual views (surface and section) were evaluated on two different endoscopic sets obtaining an approximate accuracy of 84\%. This work demonstrates that proper initialization of weights (from other datasets) and learning with few samples to refine the model is effective to generalize the classification of the images with a promising performance on individual views. 
Although the results are promising, the effect can only be measured on individual views, which differs with the way visual inspection is performed in MCA.

The authors in (\cite{villalvazo2022improved}) implemented a DL-based model assisted by a multiview approach in order to fuse information from surface and section views. This implementation has demonstrated the potential of embeddings to represent relevant information useful for the classification stage.

This paper revisits the use of metric learning for kidney stone classification, albeit with some extensions inspired by recent development in the state-of-the-art. The reason for choosing a DML approach rather than other types of methods derives from its notable capability to handle image classification tasks like face recognition (\cite{ranjan2017l2, liu2017sphereface, wang2018cosface, deng2019arcface}). For instance, tasks where different objects with various backgrounds, poses, and illuminations are present might mislead traditional DL classification models and are thus more amenable to DML methods. 

While satisfactory results can be achieved with other classification DL models (e.g., Convolutional Neural Networks) that can take these factors into account, it regularly demands to have a great amount of available data. DML methods have shown noticeably impressive results for many other tasks, such as anomaly detection, image retrieval, person re-identification, etc. (\cite{kaya2019deep}). 

One aim of this paper is to show that DML-based approaches have the capacity to successfully cope with the challenging conditions found in endoscopic imaging (i.e., variable illumination conditions, and blurred images, among other factors) and can be used for implementing reliable models for kidney stones identification, with improved performance when compared to other models in the state-of-the-art based either on SML methods or DL architectures.
\subsection{Deep Metric Learning \label{DeepMetricLearning}}

In the DML literature, there are two elemental approaches to the input data used by the model: pairwise or triplet inputs. Among the pairwise input models, one of the most popular is the \textit{Siamese Network} which gathers various losses. One of the most fundamental is the \textit{contrastive loss} (\cite{chopra2005learning, hadsell2006dimensionality}), where small positive pairwise distances and negative pairwise distances above a certain margin are encouraged.
Concerning the triple input models, the \textit{triplet network} is one of the most remarkable. Among the cost functions within the triplet sample paradigm, the \textit{triplet margin loss} (\cite{hoffer2015deep}) is among the most popular. This loss function considers three types of samples: positive, negative, and an anchor. Here it is intended that the distance between the anchor and negative samples should be greater than the distance between the anchor and positive samples by at least a given margin. The triplet margin loss has an advantage over the contrastive loss: it accounts for the intra-class and inter-class relationships (\cite{musgrave2020metric}). The use of the variance differences between classes makes this method less restrictive.

For each approach, there is a variety of available loss functions, e.g., Angular loss (\cite{wang2017deep}), Mixed loss (\cite{chen2018dress}), Margin Loss (\cite{wu2017sampling}), Multi-similarity loss (\cite{wang2019multi}), N-Pairs loss (\cite{sohn2016improved}), and others (\cite{deng2019arcface}). However, the use of paired or triplet samples usually involve a high computational and memory cost, impacting the training process. \textit{Sample mining} strategies can be used to identify the most informative samples, capable of increasing the performance, as well as reducing the training speed. Nevertheless, in the case of hard-negative mining for instance, it is the strategy with the contrastive loss which generally converges faster. Moreover, the use of the triplet loss often leads to noisy gradients and collapsed embeddings, i.e., all samples have the same embedding (\cite{musgrave2020metric}). 

On the other hand, semi-hard negative mining is recommended for triplet loss over hard-negative mining to avoid the risk of \textit{overfitting}. However, in some cases, it might converge quickly at first, but as the number of negative samples within the margin runs out, it drastically slows down its progress toward the objective function minimization (\cite{wu2017sampling}). This implies that choosing an appropriate sampling strategy could be a difficult decision, as it is too sensitive to the properties of the underlying dataset or to changes in the used architecture.

Nevertheless, new approaches have extended DML methods to new research areas, showing improvements in leveraging data relationships. For instance, the Zero-Shot Learning (ZSL) and Few-Shot Learning (FSL) paradigms have been proposed to address applications for which a limited number of samples are available. Such approaches overcome the small dataset issue through the use of prior knowledge to generalize faster, even to other domains (\cite{wang2020generalizing,milbich2021characterizing}). The ZSL approach, in which train and test class sets are disjoint, aims at learning representation spaces that capture and transfer visual similarity to unseen classes (\cite{milbich2021characterizing,roth2021simultaneous,brattoli2020rethinking}). The ZSL approaches face the challenge of constructing a priori unknown test distributions with an unspecified distributional shift from the train data. However, arbitrarily large distributional shifts may cause the captured knowledge from the training data to be less significant to the test data (\cite{milbich2021characterizing}), leading thus to an ill-posed learned representations problem.
FSL approaches, for which at least few samples of the test distribution are available during training, improve the quality of the embedding representations (\cite{tian2020rethinking,rajasegaran2020self,sung2018learning,snell2017prototypical}). Specifically, in \cite{milbich2021characterizing} it has been proven that adaptations of FSL can improve the generalization capabilities of DML since even the minor additional domain knowledge provided helps to adjust the learned representation space to achieve better Out-Of-Distribution (OOD) generalization, i.e., due to covariate shift. 

This contribution is based on GEMINI, a guided (i.e., a teacher-student scheme approach) deep metric learning approach that exploits lessons from FSL approaches to improve the generalization capabilities of downstream classification tasks. It is shown how such a method can be effectively used for implementing a kidney stone image classification algorithm with superior performance to other DML methods in the state-of-the-art, while maintaining or surpassing the performance of recent DL-based methods, due to the use of feature fusion strategies.

\subsection{Paper structure \label{PaperContent}}
The rest of this paper is organized as follows. Section \ref{proposed-solution} provides an overview of the proposed DL-architecture and  model adaptations. Section \ref{materials-methods} presents the dataset, model parameter adjustment and quality criteria. Section \ref{experiments} compares the results of the surface and section images evaluated with the proposed method on individual views. In addition, Section \ref{experiments} describes the results of fusion strategies for combined views. Section \ref{discussion} discusses the results. Finally, Section \ref{conclusion} concludes this contribution and proposes perspectives.
\section{Overview of the proposed DL-architecture \label{proposed-solution}}

The solution proposed in this contribution is a DML-based approach, referred to as \textit{Guided Deep Metric Learning} (GDML). Unlike numerous other DML approaches, GDML consists of two phases. First, a \say{teacher model} produces a source embedding space (reduced hypothesis space) that reduces the complexity of the learning task, an alternative to using sample mining strategies.
Thus a \say{student model} is trained over the generated reduced hypothesis space that helps to prevent overfitting and improves the student model's ability to generalize to new unseen data. 

\subsection{Components of the DL-architecture}
\label{ComponentsOfTheArchitecture}
\begin{figure}[tb]
\centering
\captionsetup{justification=justified}
\includegraphics[width=.48 \textwidth]{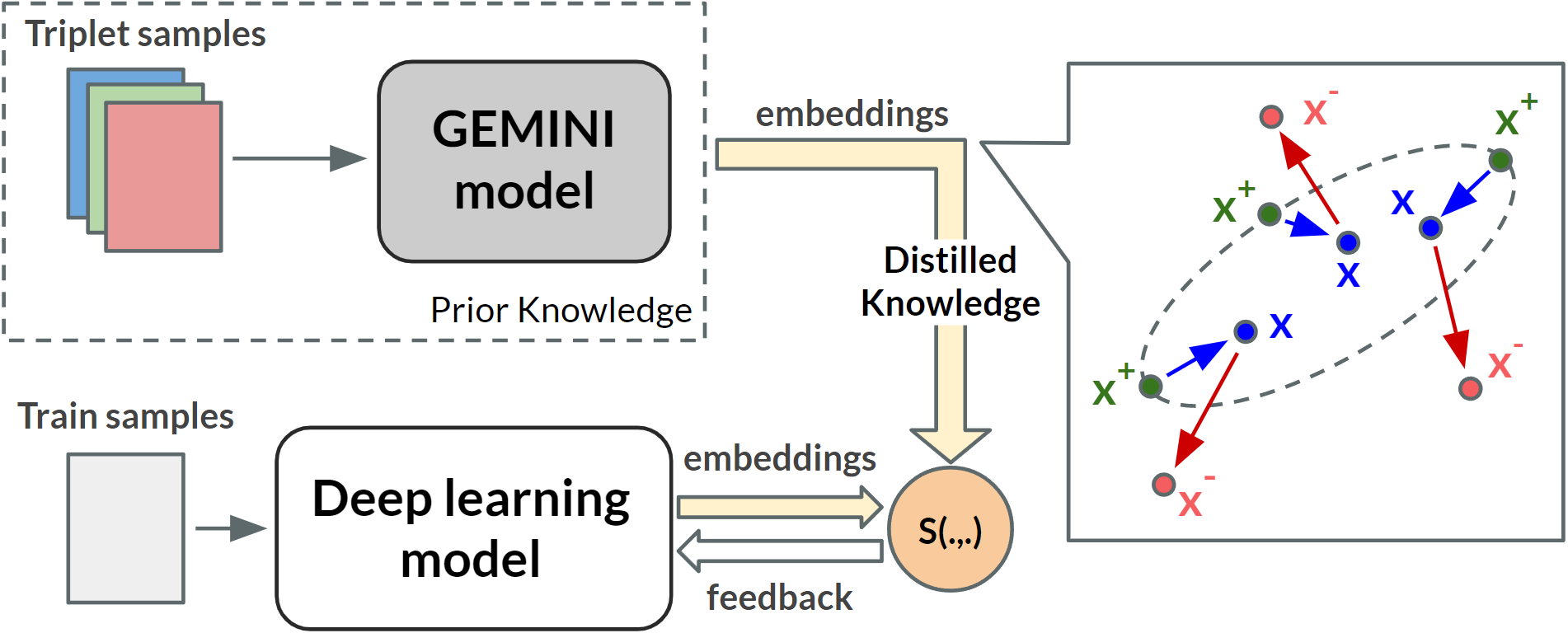}
\caption{Schematic representation of the proposed architecture. The reduced hypothesis space generated in an a priori-way by the GEMINI model is used to guide the training of the DL-model using the teacher-student architecture.}
\label{fig:propBlocks}
\end{figure}

The proposed DL-architecture consists of two independent models and exploits key concepts on both FSL and Knowledge distillation.
The first, its a master model, referred to as \textit{GEMINI}. It is a multi-branch model which generates a reduced hypothesis (a priori) space . The second, is a student DL-model trained to learn an embedding function which is compared to the feature space produced by GEMINI through a similarity function (see Fig.~\ref{fig:propBlocks}).

These two individual components of the proposed architecture are detailed in the following sections. 

\subsection{GEMINI Model}

\begin{figure}[t]
\centering
\captionsetup{justification=justified}
\includegraphics[width=.45 \textwidth]{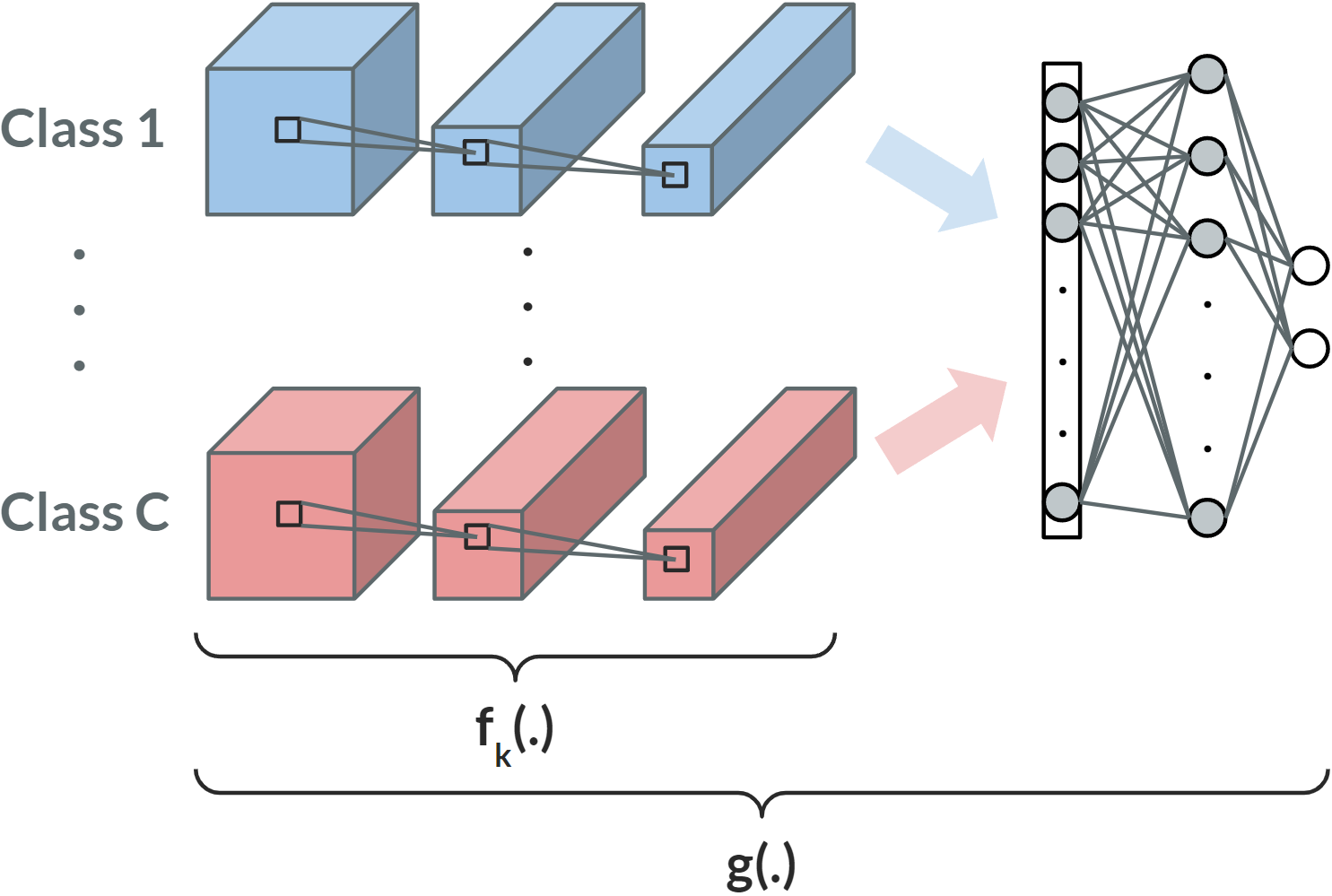}
\caption{Diagram of the GEMINI model architecture.}
\label{fig:geminialone}
\end{figure}

Analogous to parameter sharing strategies used in FSL (\cite{wang2020generalizing}), the GEMINI model consists of two components (see Fig.~\ref{fig:geminialone}). The first component $f_k(\cdot)$ is in charge of exploiting the local information of the different classes, each class being associated with a ``one stream layer'' (\cite{aghamaleki2019multi,chenarlogh2019multi}). Then, the global fully connected component  $g(\cdot)$ attempts to exploit the information of the local representations by sharing parameters between the different classes.  This approach enables to avoid the strong restrictions of a classification layer, e.g., the use of cross-entropy.

The model uses a triplet dataset $X$ generated from a training dataset $D_{train}$ with $c$ classes. The triplet data set considers three types of samples: an anchor sample $\vx_{(k)}$, a positive sample $\vx^{+}_{(k)}$ and a negative sample $\vx^{-}_{(l)}$.
The samples $\vx_{(k)}$ and $\vx^{+}_{(k)}$ are considered to be similar (i.e., they belong to a same class $k$), while the negative sample $\vx^{-}_{(l)}$ is considered as dissimilar to the anchor and positive samples (class $l$ $\neq$ class  $k$).  Each sample is fed to the network through its respective stream, depending on its class. The outputs are intermediate representations denoted by $f_i(\vx^{*}_{(i)})$ for each class $i = 1 \dots c$, regardless of the sample type (anchor, positive or negative).

Then, once the samples have passed through the local component $f_k(\cdot)$, they are  all fed to the global component $g(\cdot)$ where the samples in the mini-batch share the layer parameters. The embedded representation of the network is denoted by $g(\vx^{*}_{(i)}) = g(f_i(\vx^{*}_{(i)}))$. Thus, once the intermediate and final representations are obtained for each sample in the triplet sample, both components of the network are coupled through the cost function given in Eq. (\ref{costfunc}) 

\noindent
\begin{equation}
\begin{array}{l}
L\left(f, g, d, \beta, M\right) =  \hfill\\ 
  \qquad \qquad \hspace*{-16mm} \displaystyle \sum_{\left(\vx_{(k)},\vx^{+}_{(k)},\vx^{-}_{(l)}\right) \in X_{b}} \beta \cdot d\left( f_k \left(\vx_{(k)}\right), f_k \left(\vx^{+}_{(k)}\right)\right) \quad +  \\ 
 \quad\qquad \hspace*{3mm} \displaystyle (1-\beta) \cdot \left[ M - d\left(g\left(\vx_{(k)}\right), g\left(\vx^{-}_{(l)}\right)\right)\right]_{+}  \label{costfunc}
\end{array} 
\end{equation} 

where $\beta \in \left[0,1\right]$ is a weighting parameter, $d$ corresponds to a distance function (an Euclidean distance here), $[a]_{+}=max(0,a)$ stands for the hinge loss, and $M$ is a margin. 
The first part of the cost function focus on local information of the class and evaluates the closeness of similar samples; it emphasizes the proximity of the intermediate representations of the same class samples. The second part of the cost function has been added to evaluate the distanciation of samples of different classes; it penalizes smaller distances than a margin between samples.
Thus, the first term in the sum in Eq. (\ref{costfunc}) minimizes intra-class distances, and the second term in the sum prevents trivial solutions by maximizing inter-class distances. The $\beta$ weight balances the impact of both terms in the loss function.

Term $M$ is defined by $M = d(g(\vx_{(k)}), g(\vx^{+}_{(k)})) + m$, where $m$ is a margin. 
This makes the second part of the cost function to resemble the triplet loss function, but with an additional term that keeps the similar samples together using the local information.

The model has two simultaneous effects: it decreases the distances between anchor and positive samples, while it also increases the distance between anchor and negative samples (see Fig.~\ref{fig:propBlocks}). Sample mining techniques can be avoided due to the use of the GEMINI master model.  
As mentioned in Section \ref{DeepMetricLearning}, such techniques have been proven not to work well in all scenarios and to restrict generalization capabilities of the model.

\subsection{Complete Architecture} 

In this contribution, the complete architecture is based on an offline knowledge distillation approach in which different DL-models can act as student network.  A PyTorch ResNet-50 implementation was used here for the student network. 
The advantage of this architecture is that GEMINI has already searched through the space, and it has arrived at a reduced hypothesis space a priori. In this way, the ResNet model is expected to need fewer samples to converge to a suitable hypothesis closer to the searched optimum. With this approach the risk of overfitting is also reduced (\cite{wang2020generalizing}). Both models (i.e., the GEMINI teacher and ResNet student model) are coupled by a similarity function, $s(\cdot,\cdot)$, which measures the deviation of the ResNet hypothesis from the reduced hypothesis space.

\begin{figure*}[thb]
\centering
\captionsetup{justification=justified}
\includegraphics[width=.99 \textwidth]{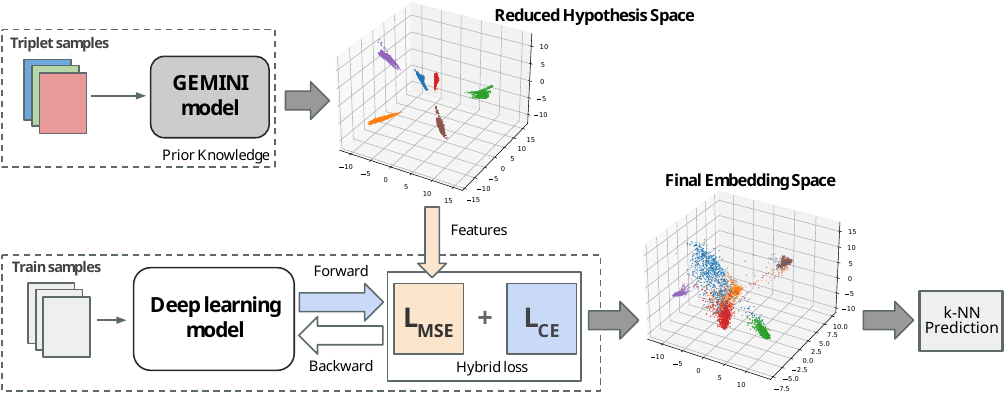}
\caption{Block diagram of the adaptations made for the GEMINI model. Example of a reduced hypothesis space generated a priori by the GEMINI model to which the output embeddings of the student model will be compared. In addition, a classification layer is added to the student model, making use of an hybrid loss function to improve performance.}

\label{fig:Adapmodel}
\end{figure*}

\subsection{Model adaptations}
Some adaptations have been made to the ResNet student model to improve the global architecture performance. The output of the student model has been extended by adding a classification layer with a fixed output size (number of classes) and measuring the classification capacity through the Cross-Entropy loss function.
Consequently, a hybrid loss function is used for the student model, where the distances between the representations (embeddings) and the classification capacity in the model output are considered for performing the final prediction (see Fig.~\ref{fig:Adapmodel}).

In this contribution, the similarity function, $s(\cdot,\cdot)$, which measures the deviation between the GEMINI teacher model and the ResNet student model, is reformulated as a hybrid loss function, where both losses together balance the invariance and discrimination capacity of the model.


\begin{equation}
\label{costcop}
S\left(\gamma, Z, \hat{Z}\right) = \sum_{i=1}^{|B|}  \left( \gamma \cdot d\left(\vz_{i}, \hat{\vz}_{i}\right) + CE\left(\vy_{i}, \hat{\vy}_{i}\right)\right)
\end{equation}

\noindent
In the proposed hybrid loss function (see Eq.~(\ref{costcop})), the term $d(\cdot,\cdot)$ is a distance function (Euclidean distance), and $CE$ represents the Cross-Entropy loss function. The term $\gamma \in \left(0,1\right)$ is a compensation parameter that adjusts the inequality between the two different cost functions that take as input the set of representations $Z = \{\vz_i,y_i\}^N_{i=1}$ of the student model and the embeddings $\hat{Z} = \{\hat{\vz}_i,\hat{\vy}_i\}^N_{i=1}$ generated by the teacher model, both in the training batch $B$.\\

\section{Experimental Setup }
\label{materials-methods}
%
The performance assessment of the proposed architecture is based on classification results obtained for surface (SUR) and section (SEC) views taken each separately, as well as by evaluating DL-networks which jointly exploits the two views (i.e., through an image fusion-like scheme). 
The collected dataset is presented in subsection \ref{dataset}. Then, this section describes how different models were trained in order to assess the performance of section, surface, and fusion  models. In particular, various ways of mixing the information (surface and section views), are explored to carry out a kidney stone identification using visual information seen in the images of both views.   

\subsection{Ex-vivo endoscopic dataset}
\label{dataset}

\begin{figure*} [t] 
    \centering
    \subfloat[SUR ]{
    \label{fig:dataseta}\includegraphics[width=0.485\linewidth]{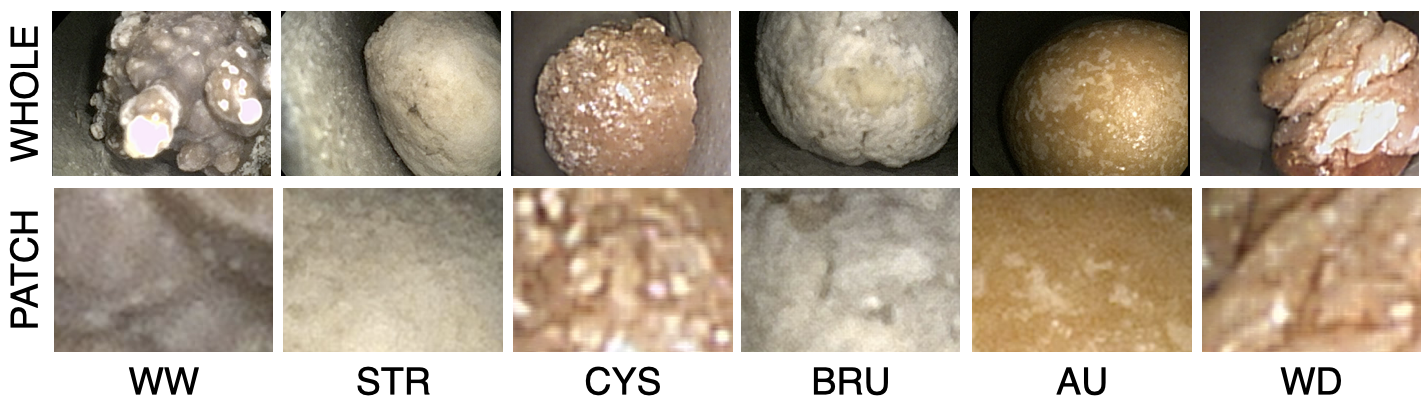}}
    \hspace{0.1cm}
    \subfloat[SEC]{
    \label{fig:datasetb}\includegraphics[width=0.485\linewidth]{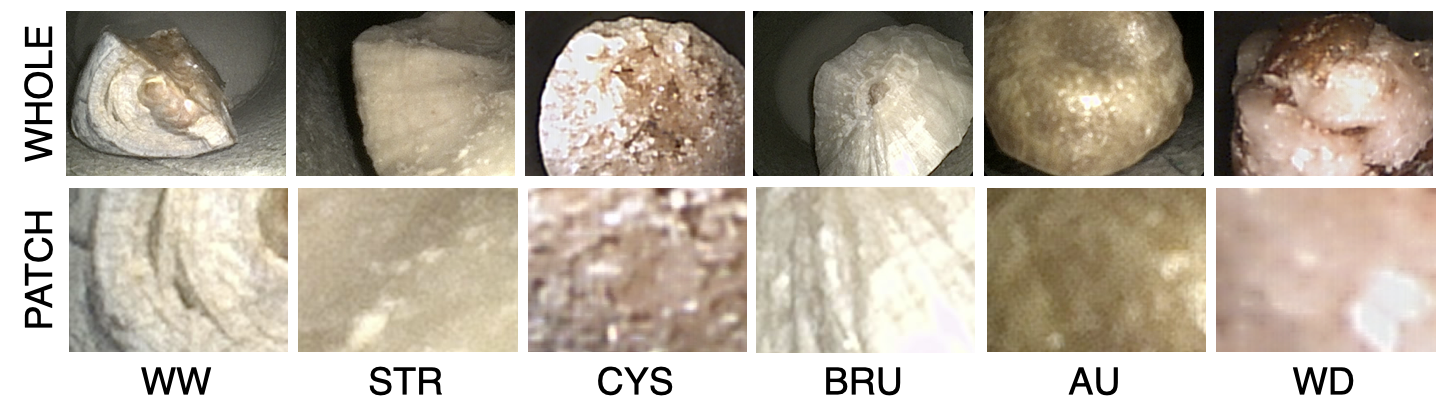}}
    \vspace{-0.1cm}
    \caption{Examples of the six most common kidney stone types: Whewellite (WW), Struvite (STR), Cystine (CYS), Brushite (BRU), Uric Acid (UA), and  Weddellite (WD). The complete  kidney stone images are given by the upper row in sub-figures (a) and (b) respectively for the surface (SUR) and section (SEC) views of the fragments. Patches extracted from the images are in the lower row.}
    \label{fig:dataset}
\end{figure*}

The endoscopic images used in this contribution (see Fig. \ref{fig:dataset} and Table \ref{tab:dataset}) were acquired by an urologist and their type was identified in a biology laboratory using the reference MCA  procedure described Section \ref{sec:MedicalContext}.
The dataset consists of 246 surface and 163 section images, for a total of 409 images. The images  represent six of the most common kidney stone types (\cite{corrales2021classification}) and are sorted by subtypes denoted by WW (Whewellite, subtype Ia),  WD (Weddellite, subtype IIa),  UA (Uric Acid, subtype IIIa), STR (Struvite, subtype IVc),  BRU (Brushite, subtype IVd) and CYS (Cystine, subtype Va). 

The images  were captured in ex-vivo by placing  kidney stone fragments inside a tubular enclosure  simulating the shape, textures  and color of the inner epithelial walls of ureters (for more details, see \cite{el2022evaluation}). These images are visually close to those acquired in in-vivo (during an ureteroscopy) since the fragments were acquired with an ureteroscope (the endoscope actually used during an ureteroscopy and with its standard image quality) and by simulating realistic clinical in-vivo conditions (the scene and it's illumination by an endoscope).
%
%

\begin{table}[t]
\centering
\caption{Number of endoscopic images sorted in  two sets: the set of surface (SUR) views  and the set of section (SEC) views. MIX refers to as the sum of the SEC and SUR images gathered  in a third image set.}
\label{tab:dataset}
\begin{tabular}{@{}ccccc@{}}
\toprule
Main component    & Subtype    & SUR & SEC & MIX \\ \midrule
Whewellite (WW)        & Ia          & 62  & 25  & 87  \\
Weddellite (WD)        & IIa         & 13  & 12  & 25  \\
Uric Acid (UA)      & IIIa        & 58  & 50  & 108 \\
Struvite (STR)         & IVc         & 43  & 24  & 67  \\
Brushite (BRU)         & IVd         & 23  & 4   & 27  \\
Cystine (CYS)          & Va          & 47  & 48  & 95  \\ \midrule
\multicolumn{2}{r}{Total images per view} & 246 & 163 & 409 \\ \bottomrule
\end{tabular}
\end{table}

Image sub-parts were used for the kidney stone type recognition rather than the complete color frames which have a size of 1920$\times$1080 pixels (see the upper row in Fig.~\ref{fig:datasetb}). As in all previous works on kidney stone classification, patches (see the lower row in Fig. \ref{fig:datasetb}) were extracted from the images for several reasons. First, the amount of images (409 in all) is rather moderate and some classes include few samples. Second, only about 15\% of the pixels effectively correspond to kidney stone fragments, the remaining pixels represent the scene background. Third, as seen in Table \ref{tab:dataset}, the six classes are imbalanced in terms of samples. Extracting patches can partly compensate the drawbacks related to the available image dataset.   

The work in (\cite{ochoa2022vivo}) has shown that square patches with an appropriate size (i.e., $256\times256$ pixels) are able to capture sufficient color and texture information to identify the type of kidney stones. 
%
The patches are shifted over the images, but a patch is only extracted from an image when it has a maximal overlap of 20 columns and rows with the patches already extracted from the same frame. This overlap limitation allows to avoid the extraction of redundant information within an image of the same kidney stone fragment. 
%
%
%
%
As in (\cite{lopez2021assessing}), the patches $P_i$ extracted from images $I_i$ were ``whitened'' using the mean $m_i$ and standard deviation $\sigma_i$ determined in each channel of the color images ($i = R$, $G$ or $B$). The whitened patch values are given by  $I_i^w = (I_i - m_i)/\sigma_i$. 

The partitioning strategy for the training and test set was $80\%$ and $20\%$, respectively. A random and non-repeating dataset partitioning strategy was used to avoid data leakage in the datasets  (for more details, see \cite{ochoa2022vivo}).
\subsection{Implementation and training details}
As most often done in previous works dealing with kidney stone identification, the models used in this contribution are trained two times, one time exclusively on SUR data, and another only on SEC views. However, the training on the individual SUR and SEC models (embeddings) are exploited in different fusion strategies to combine the results from both views.

The proposed  training protocol uses a ResNet-50 as backbone for all the networks,  i.e., including the student model in our proposal (the teacher model is a modified Convolutional Neural Networks). All models were implemented using Python 3.8 and Pytorch 1.6., and the training was performed on  Nvidia V100, P100, and T4 GPUs with no more than 16 Gb memory available. 

The training process of the models ran for no more than 60 epochs. Neither data augmentation, nor additional pre-processing was applied to the patches. 

For all networks, different sizes in powers of 2 were tested for the output embeddings (i.e., 8, 16, 32, 64, 128, 256, 512, 1024). The batches were  randomly constructed, assuming a uniform distribution. Finally, the results were averaged over multiple random initialized seeds.

The repository by (\cite{musgrave2020pytorch}) was used for some losses, reducers, sample mining functions, and DML metrics. For this study, following evaluation metrics were used to quantify the kidney stone type recognition performance of the models: Accuracy, Precision, Recall, and  F1-Score.
All these metrics were obtained using the $k$-Nearest Neighbors ($k$-NN) classifier of scikit-learn in the test embedding space from the ResNet-50 output. 

The next section gives the results for section and surface views individually taken, as well as those obtained with the feature fusion strategy. Apart from these quantitative results, qualitative PCA (Principal Component Analysis) visualizations are also proposed to highlight the advantages of the presented architecture in terms of generalization.

\begin{figure*}[h!]
    \centering
    \subfloat[SUR]{
    \label{fig:knnSurface}\includegraphics[width=0.49\linewidth]{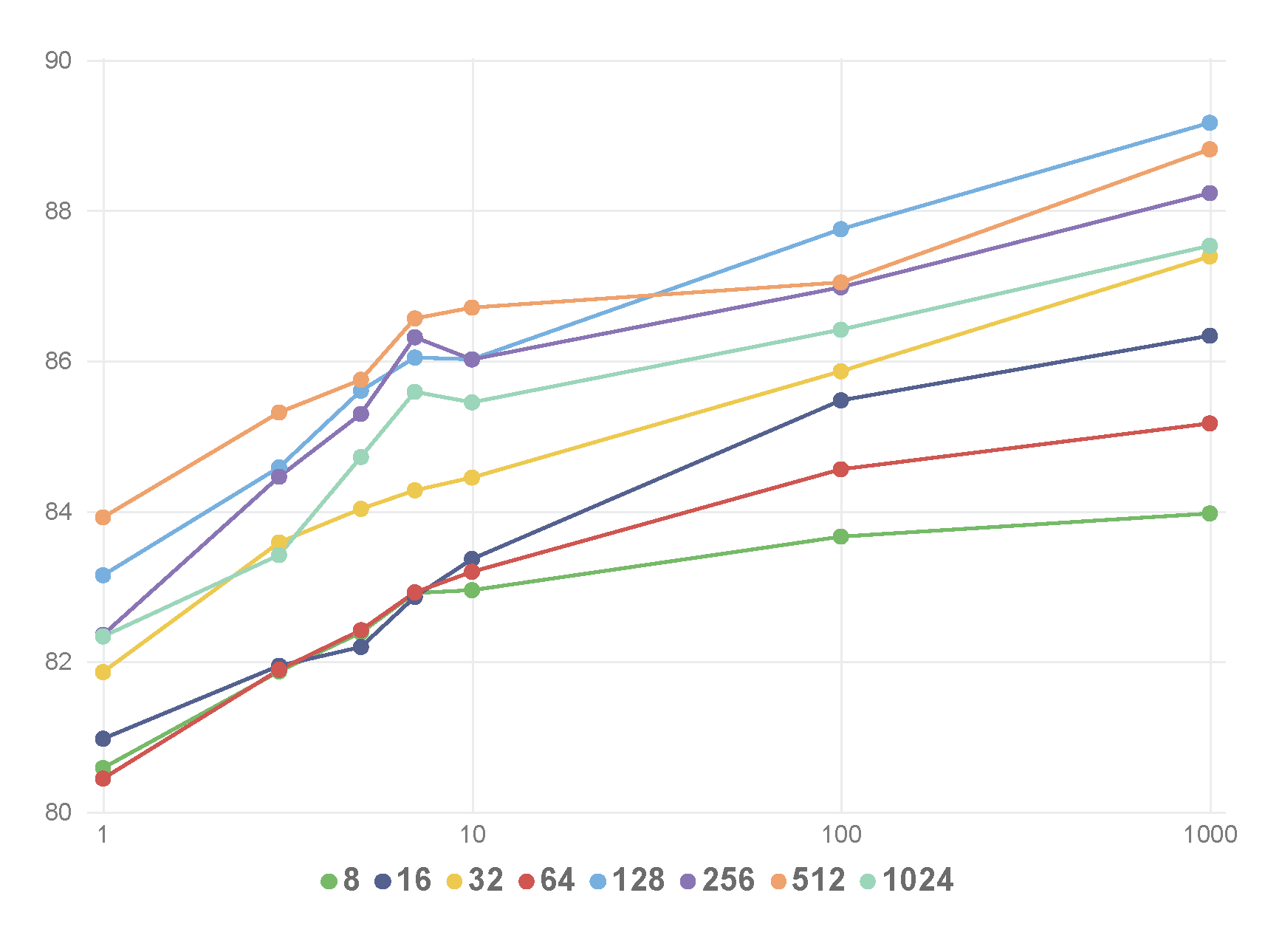}}
    \subfloat[SEC]{
    \label{fig:knnSection}\includegraphics[width=0.49\linewidth]{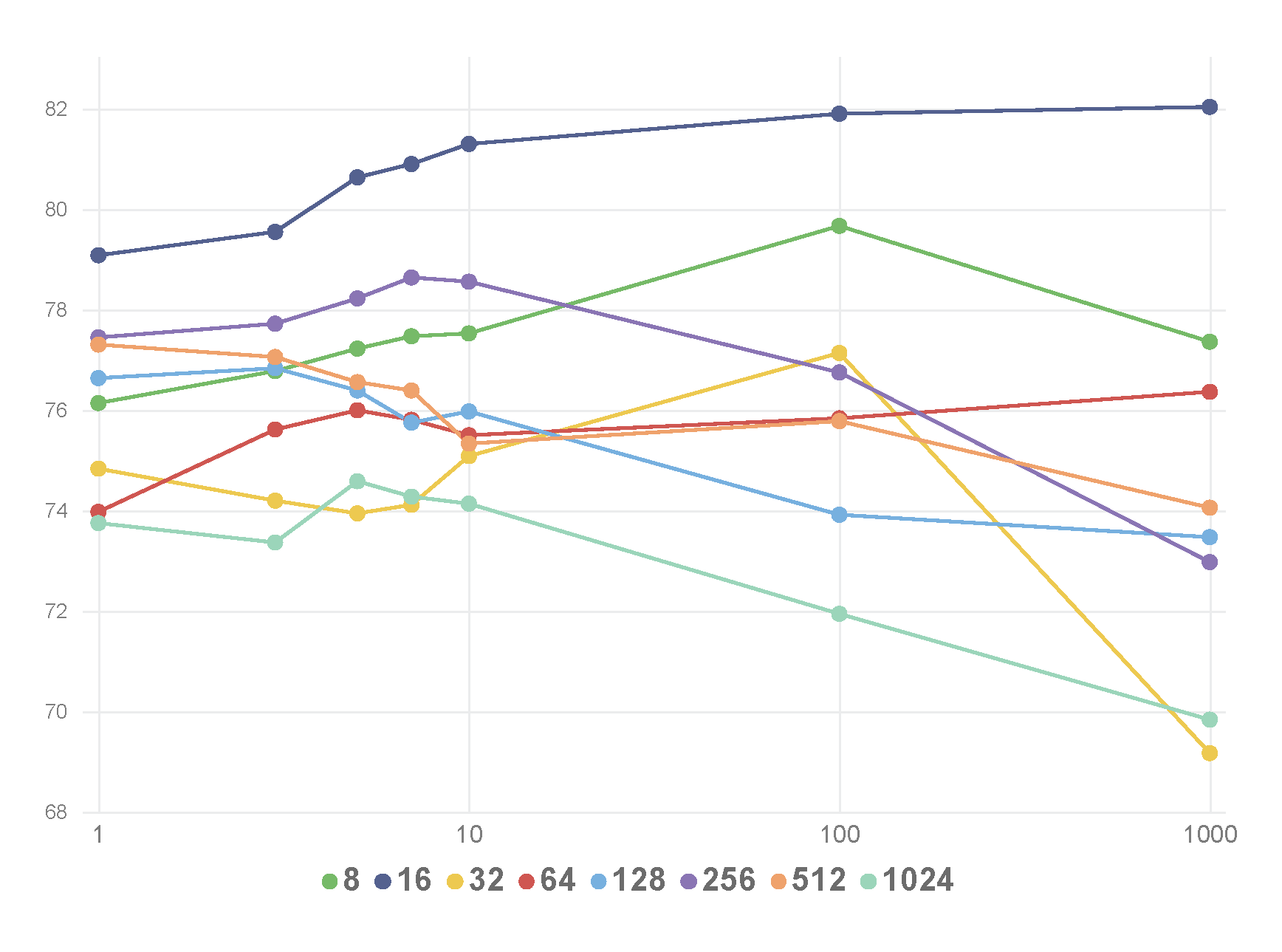}}
    \caption{Recall values (in percentage on the ordinate axes) measured on (a) SUR and (b) SEC view datasets with different embedding sizes over different values of $k$ (values on the abscissa axes) in the $k$-NN algorithm. The colors of the curves correspond to different sizes of output embeddings}
    \label{fig:knn}
\end{figure*}

\subsection{Evaluation criteria}

 The impact of the  parameter values of the ResNet-50  and GEMINI models presented in Section \ref{materials-methods} are discussed here. 
Among the performance evaluations performed to validate the  design of the proposed architecture one can mention: \textbf{i)} the influence of the number of nearest neighbors (parameter $k$ of the $k$-NN classifier), \textbf{ii)} the impact of the dimensionality of the output embedding, and \textbf{iii)} the qualitative evaluation of the dispersion/compactness of the final embedding space.

\textbf{Evaluation of the model for surface views.} 
Figure (\ref{fig:knnSurface}) allows for the analysis of the classification performace using SUR data according to  the number $k$ of nearest neighbors of the $k$-NN algorithm. It is noticeable that increasing the value of $k$ improves the performance for all the output embedding sizes. The embeddings with the middle-high dimensional configuration exhibit the best performances. 

\textbf{Evaluation of the model for sections views.} 
For the SEC dataset, one can observe a loss of performance with some configurations of embedding sizes and $k$-NN (see Fig.~\ref{fig:knnSection}). Specifically, a trend of performance loss is present for a large number of nearest neighbors in high-dimensional output embeddings.

\begin{figure*}
 \newcounter{row}
 \makeatletter
    \@addtoreset{subfigure}{row}
    
    \renewcommand\thesubfigure{\alph{subfigure}-\arabic{row}}
    \centering
    \setcounter{row}{1}%
    
    \begin{subfigure}{0.45\linewidth}
        \includegraphics[width=\linewidth]{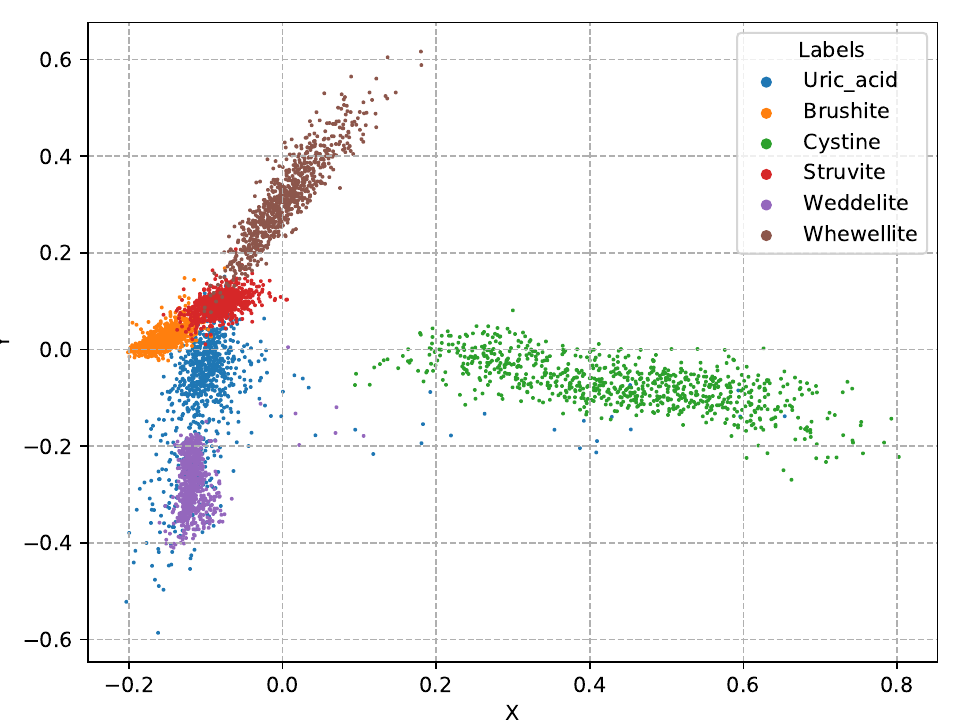 } 
        \vspace*{-5mm}
        \caption{\textbf{Siamese train embedding space (128 dimensions)}}
        \vspace*{5mm}
    \end{subfigure}\hfill
    \begin{subfigure}{0.45\linewidth}
        \includegraphics[width=\linewidth]{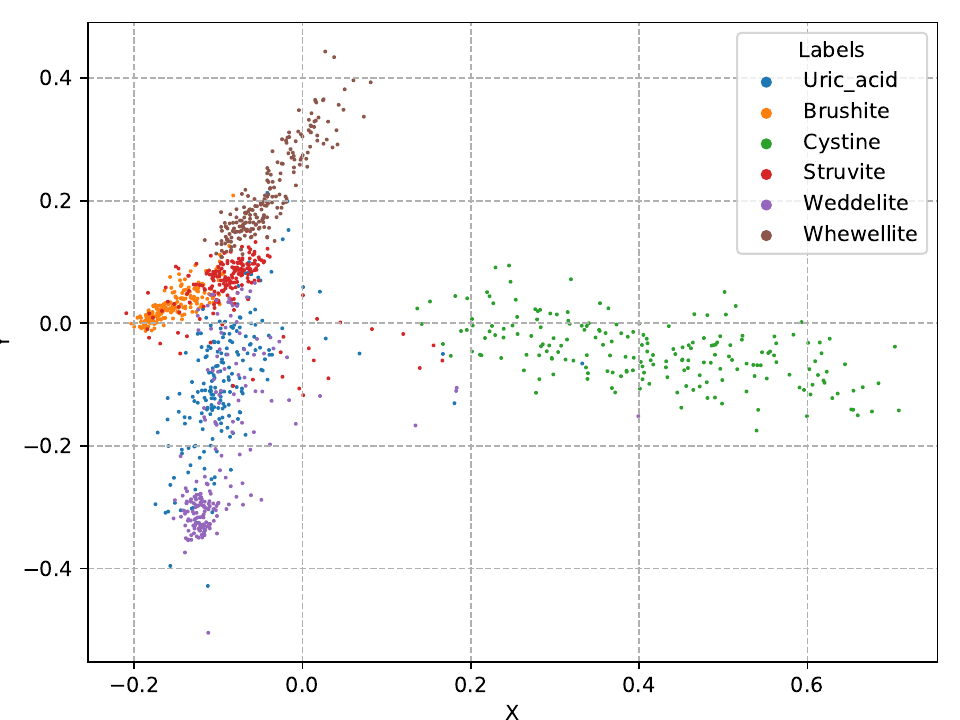}
        \vspace*{-5mm}
        \caption{\textbf{Siamese test embedding space (128 dimensions)}}
        \vspace*{5mm}
    \end{subfigure}

    \stepcounter{row}%
    \begin{subfigure}{0.45\linewidth}
        \includegraphics[width=\linewidth]{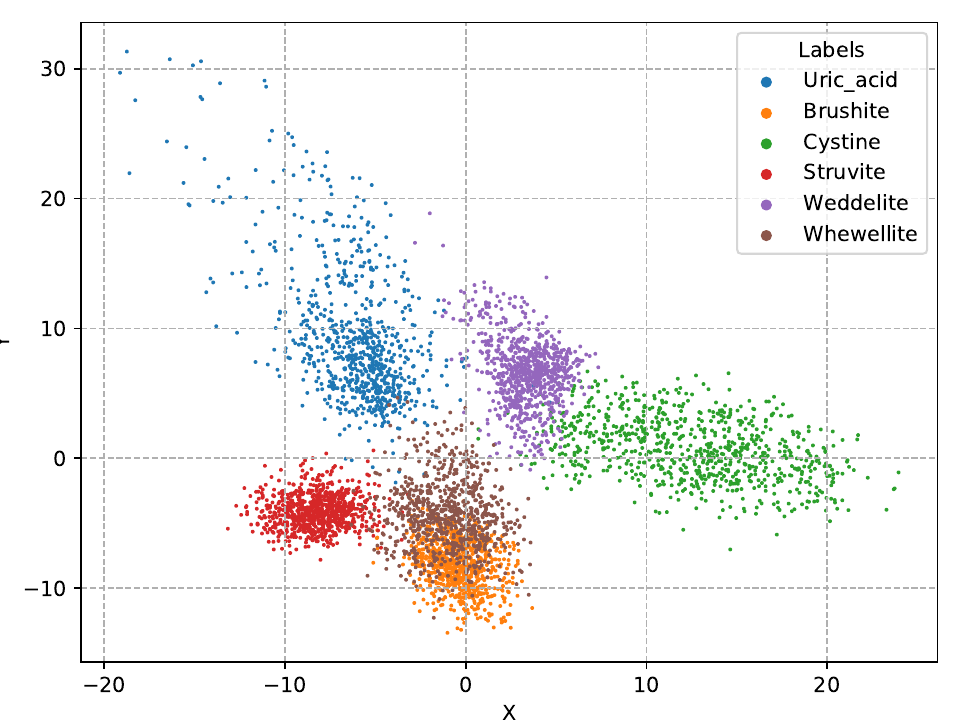}
        \vspace*{-5mm}
        \caption{\textbf{Triplet train embedding space (512 dimension)}}
        \vspace*{5mm}
    \end{subfigure}\hfill
    \begin{subfigure}{0.45\linewidth}
        \includegraphics[width=\linewidth]{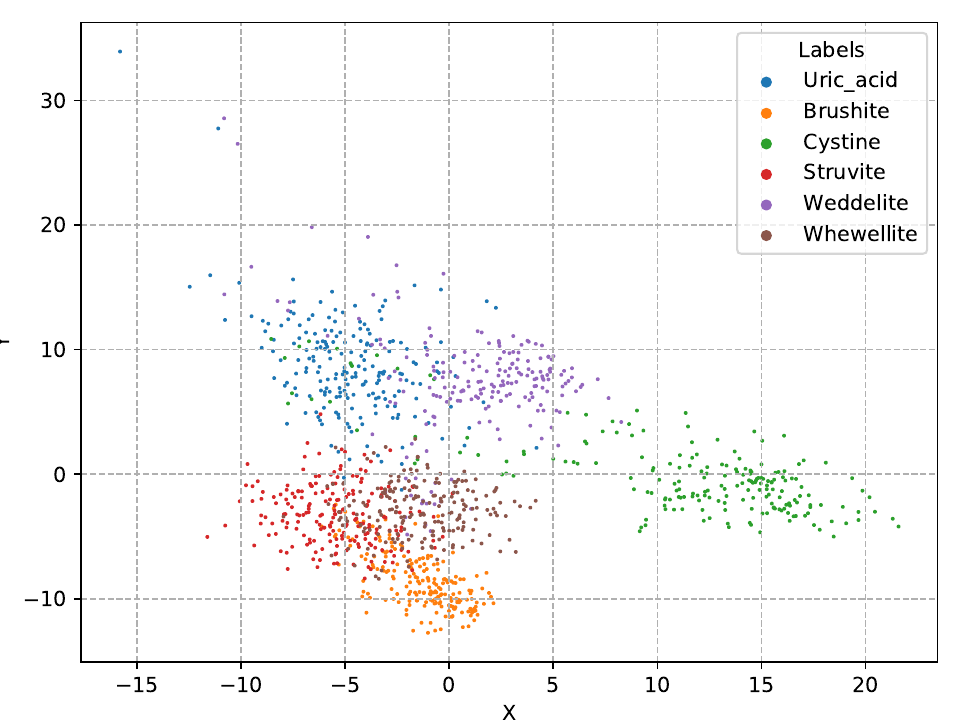} 
        \vspace*{-5mm}
        \caption{\textbf{Triplet test embedding space (512 dimension)}}
        \vspace*{5mm}
    \end{subfigure}

    \stepcounter{row}%
    \begin{subfigure}{0.45\linewidth}
        \includegraphics[width=\linewidth]{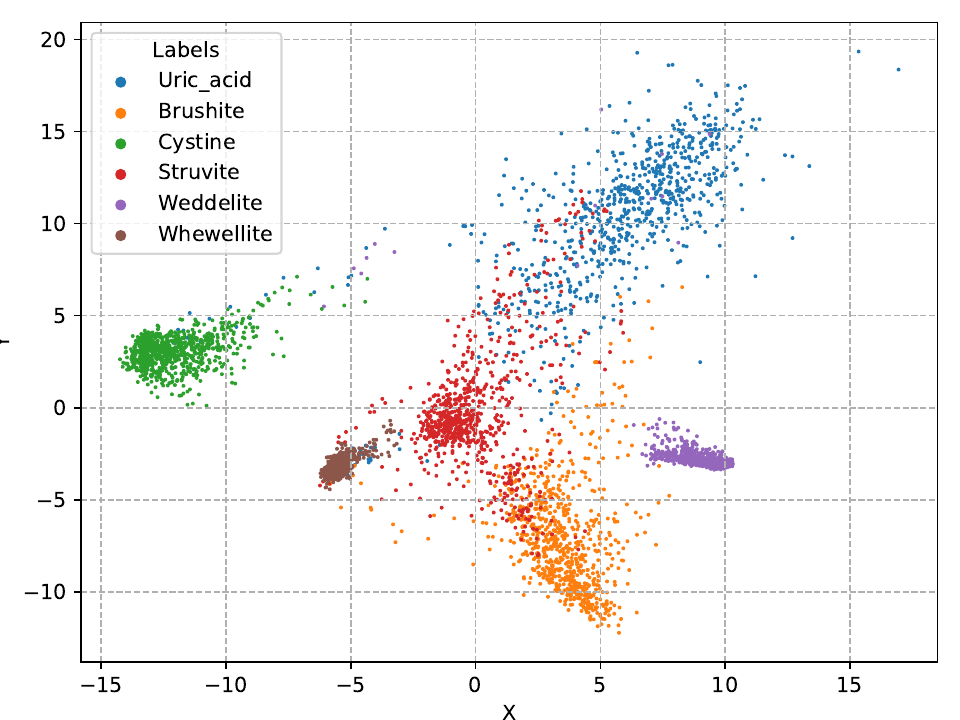}
        \vspace*{-5mm}
        \caption{\textbf{GEMINI train embedding space (128 dimensions)}} 
        \vspace*{5mm}
    \end{subfigure}\hfill
    \begin{subfigure}{0.45\linewidth}
        \includegraphics[width=\linewidth]{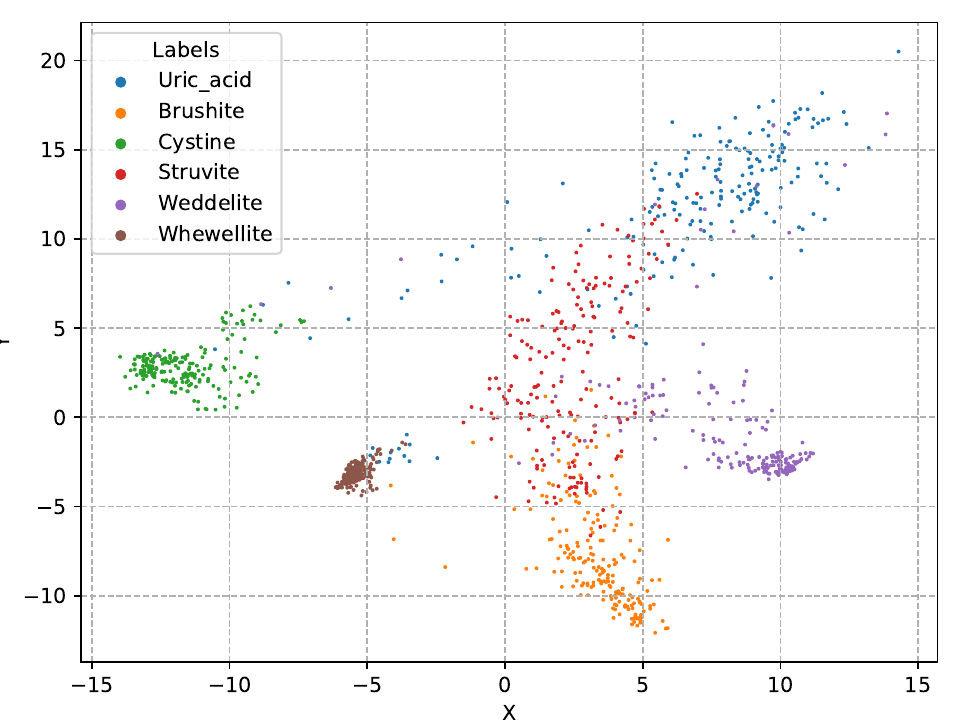}
        
        \vspace*{-5mm}
        \caption{\textbf{GEMINI test embedding space (128 dimensions)}} 
        \vspace*{5mm}
    \end{subfigure}

    \caption{DML-model embedding spaces of the surface dataset. The columns show the \textbf{a)} train and \textbf{b)} test embedding spaces from the highest performance iteration in their respective best embedding size configuration. The embedding spaces were reduced to a 2D representation using a PCA. \label{fig:DMLSUR}}
\end{figure*}

\begin{figure*}
    \newcounter{row2}
    \makeatletter
    \@addtoreset{subfigure}{row2}
    
    \renewcommand\thesubfigure{\alph{subfigure}-\arabic{row2}}
    \centering
    \setcounter{row2}{1}%
    
    \begin{subfigure}{0.45\linewidth}
        \includegraphics[width=\linewidth]{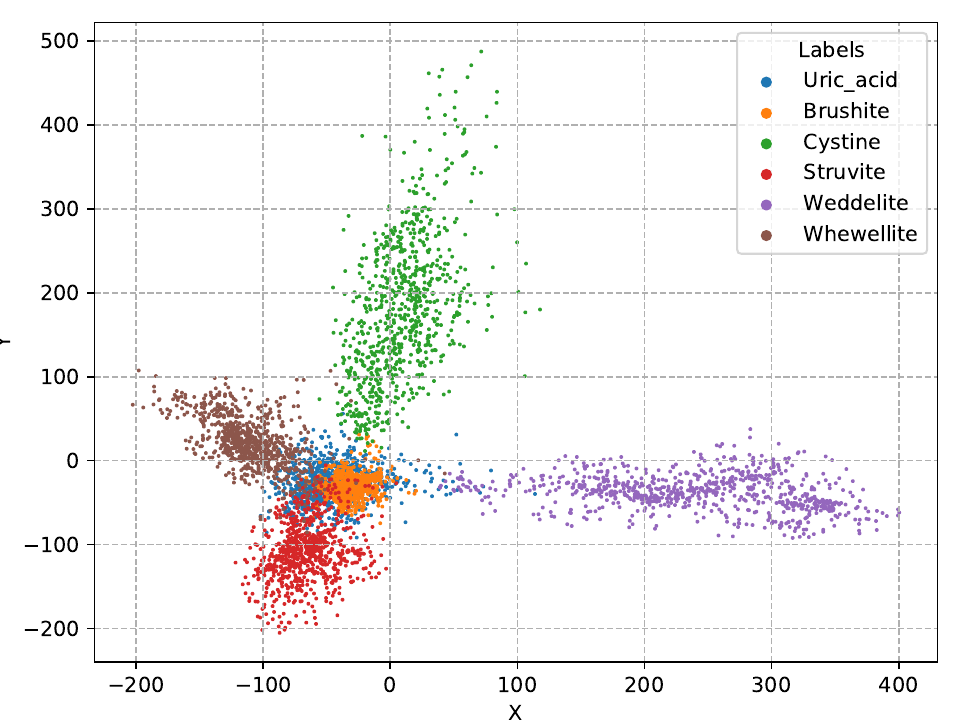}
        \vspace*{-5mm}
        \caption{\textbf{Siamese train embedding space (128 dimensions)}}
         \vspace*{5mm}
    \end{subfigure}\hfill
    \begin{subfigure}{0.45\linewidth}
        \includegraphics[width=\linewidth]{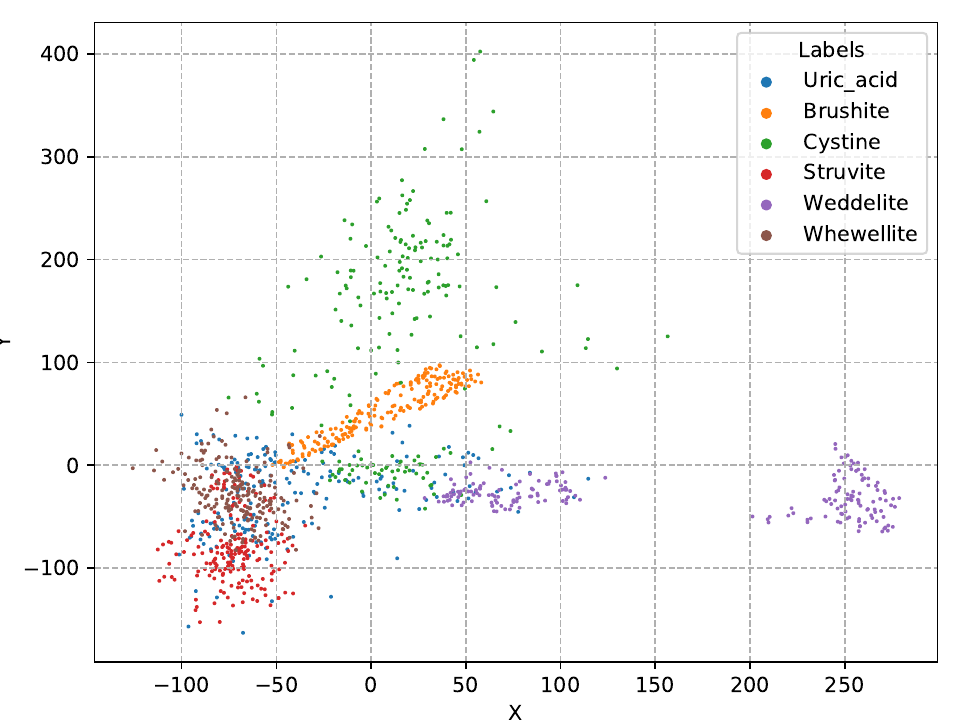}
        \vspace*{-5mm}
        \caption{\textbf{Siamese test embedding space (128 dimensions)}}
         \vspace*{5mm}
    \end{subfigure}

    \stepcounter{row2}%
    \begin{subfigure}{0.45\linewidth}
        \includegraphics[width=\linewidth]{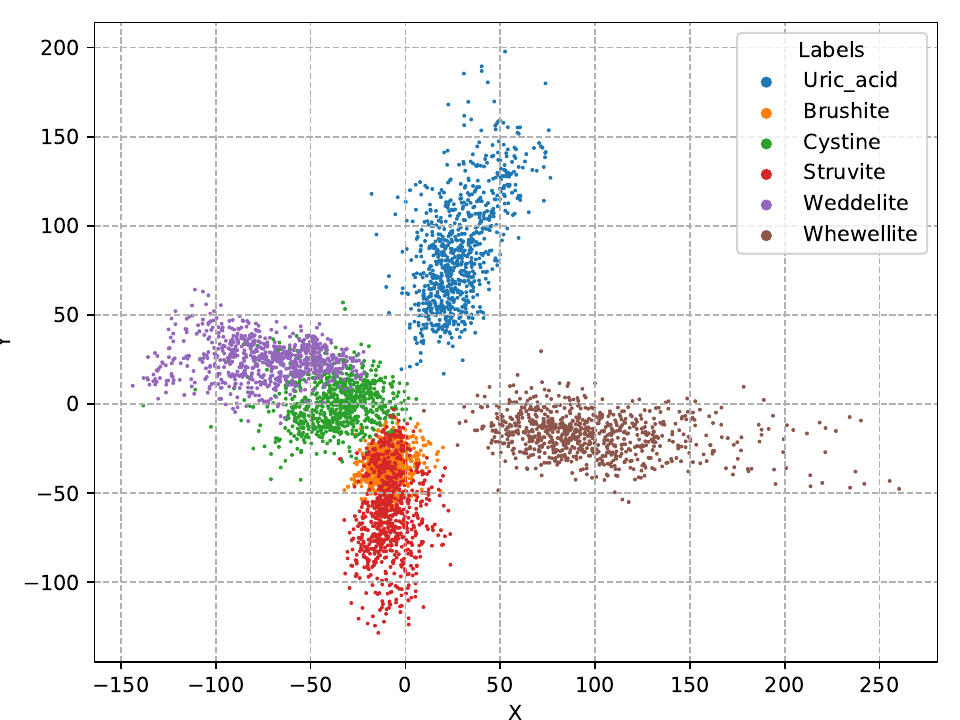}
        \vspace*{-5mm}
        \caption{\textbf{Triplet train embedding space (64 dimensions)}}
        \vspace*{5mm}
    \end{subfigure}\hfill
    \begin{subfigure}{0.45\linewidth}
        \includegraphics[width=\linewidth]{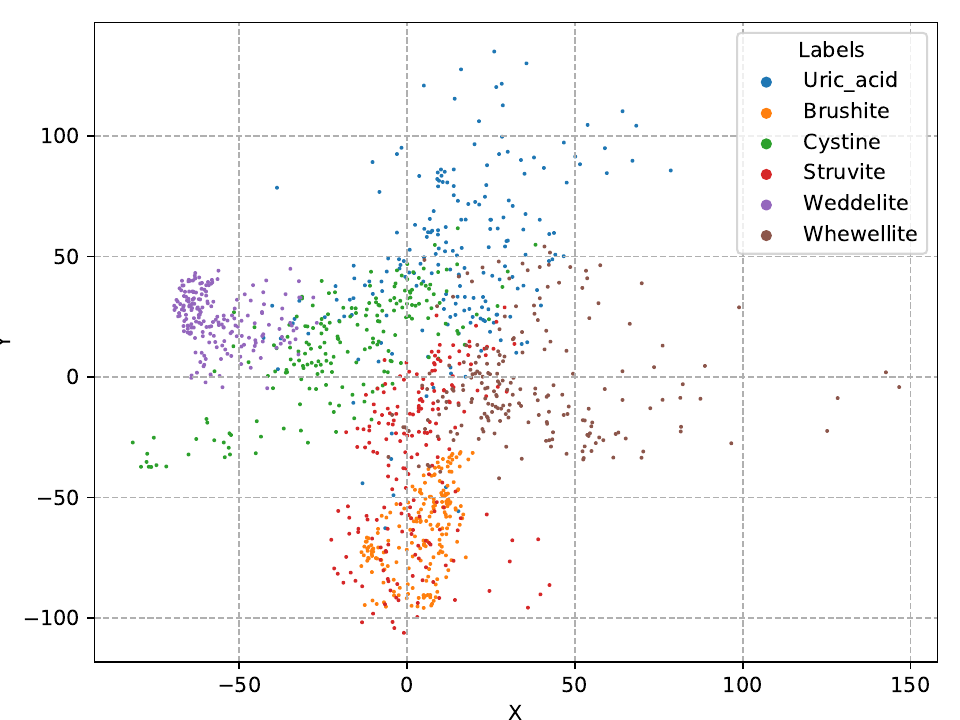} 
        \vspace*{-5mm}
        \caption{\textbf{Triplet test embedding space (64 dimension)}}
        \vspace*{5mm}
    \end{subfigure}

    \stepcounter{row2}%
    \begin{subfigure}{0.45\linewidth}

        \includegraphics[width=\linewidth]{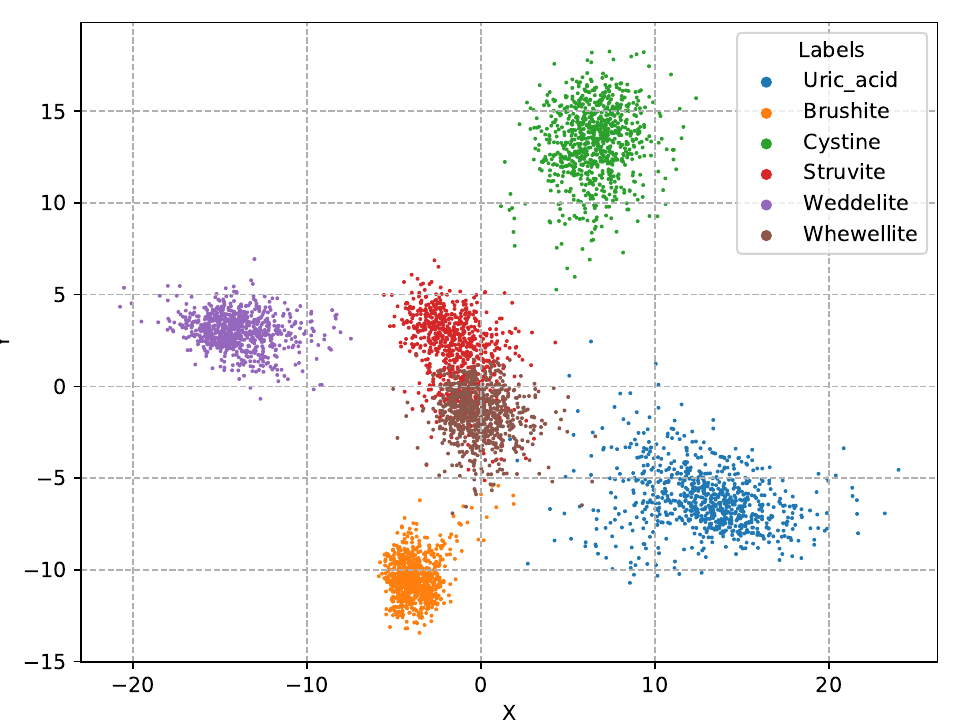}
        \vspace*{-5mm}
        \caption{\textbf{GEMINI train embedding space (16 dimensions)}}
         \vspace*{5mm}
    \end{subfigure}\hfill
    \begin{subfigure}{0.45\linewidth}
        \includegraphics[width=\linewidth]{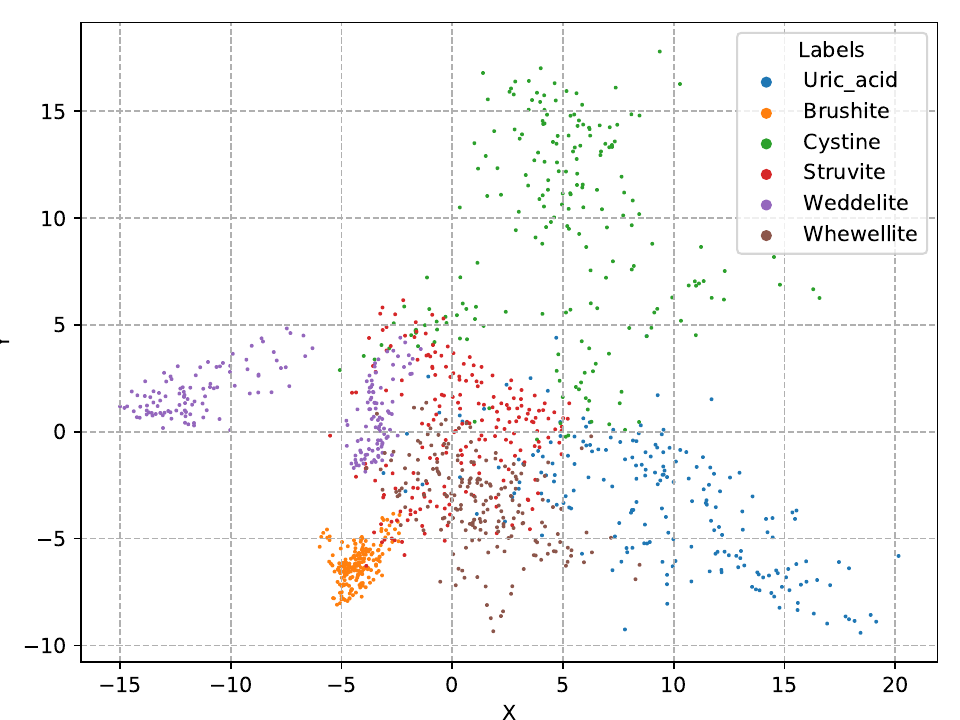}
        \vspace*{-5mm}
        \caption{\textbf{GEMINI test embedding space (16 dimensions)}}
        \vspace*{5mm}
    \end{subfigure}

    \caption{DML-model embedding spaces of the section dataset. The columns show the \textbf{a)} train and \textbf{b)} test embedding spaces from the highest performance iteration in their respective best embedding size configuration. The embedding spaces were reduced to a 2D representation using a PCA. \label{fig:DMLSEC}}
\end{figure*}

\textbf{PCA reduction evaluation.} 
The ability of the student model's solution space (i.e., the final embedding space) to be discriminating was visually assessed. To do so, a PCA dimensionality reduction was applied to the final embeddings to obtain an easily visually interpretable 2-D representation. The choice of the 2-D embedding size was motivated to push the proposed method solution to its highest potential for tasks such as data visualization or an overall understanding of the data.
Despite the loss of information introduced by the dimensionality reduction of the space, it can be  observed in Figs.~\ref{fig:DMLSUR} and \ref{fig:DMLSEC} that the model has learned a space that in general maintains a noticeable separation of the different classes. Interpretations of this figures will be provided in a later section of this paper.

The results for SUR and SEC are later individually compared with that of the state-of-the-art to show that our models exhibit a competitive performance. However, the MCA evaluation is performed using jointly the SEC and SUR views (information fusion). Several approaches have been undertaken in the state-of-the-art to determine the best fusion strategy of the SEC and SUR features. This work makes use of the fusion strategies experiments proposed in (\cite{villalvazo2022improved}) and detailed in Section (\ref{fusion-estrategies}) indor to compare the proposed model to that in the state-of-the-art of kidney stones composition identification.


\clearpage

\begin{table*}[t!]
\begin{tabular}{@{}rcccccc@{}}
\cmidrule(l){2-7}
\textbf{}                     & \multicolumn{3}{c}{\textbf{SUR}}                      & \multicolumn{3}{c}{\textbf{SEC}} \\ \midrule
\multicolumn{1}{r|}{K-NN}     & \textbf{Precision}  & \textbf{Recall}     & \multicolumn{1}{c|}{\textbf{F1-score}}   & \textbf{Precision}  & \textbf{Recall}     & \textbf{F1-score}   \\ \midrule
\multicolumn{1}{r|}{$k$ = 1}    & 85.59$\pm$0.71 & 83.14$\pm$0.39 & \multicolumn{1}{c|}{83.18$\pm$0.33} & 82.70$\pm$3.51 & 79.08$\pm$4.67 & 79.16$\pm$4.90 \\
\multicolumn{1}{r|}{$k$ = 3}    & 86.83$\pm$0.89 & 84.58$\pm$1.03 & \multicolumn{1}{c|}{84.58$\pm$0.92} & 82.77$\pm$3.33 & 79.54$\pm$3.93 & 79.67$\pm$4.07 \\
\multicolumn{1}{r|}{$k$ = 5}    & 87.45$\pm$0.56 & 85.60$\pm$0.90 & \multicolumn{1}{c|}{85.50$\pm$0.76} & 83.26$\pm$2.96 & 80.63$\pm$3.56 & 80.78$\pm$3.52 \\
\multicolumn{1}{r|}{$k$ = 7}    & 87.66$\pm$0.64 & 86.04$\pm$0.80 & \multicolumn{1}{c|}{85.86$\pm$0.70} & 83.36$\pm$3.07 & 80.89$\pm$3.68 & 81.0$\pm$3.61 \\
\multicolumn{1}{r|}{$k$ = 10}   & 87.57$\pm$0.45 & 86.02$\pm$0.77 & \multicolumn{1}{c|}{85.72$\pm$0.69} & 83.52$\pm$2.95 & 81.29$\pm$3.28 & 81.43$\pm$3.20 \\
\multicolumn{1}{r|}{$k$ = 100}  & 88.88$\pm$1.01 & 87.75$\pm$1.24 & \multicolumn{1}{c|}{87.48$\pm$1.36} & \textbf{83.86$\pm$3.26} & 81.90$\pm$4.11 & 82.15$\pm$3.91 \\
\multicolumn{1}{r|}{$k$ = 1000} & \textbf{89.81$\pm$0.89} & \textbf{89.16$\pm$0.83} & \multicolumn{1}{c|}{\textbf{88.99$\pm$0.86}} & 83.07$\pm$4.07 & \textbf{82.03$\pm$4.77} & \textbf{82.20$\pm$4.56} \\ \bottomrule
\end{tabular}
\caption{Best average performance on surface (SUR, 128-size embedding)  and  section (SEC, 16-size embedding) datasets for different values of $k$ in a k-NN classifier.}
\label{tab:comparison-stateoftheart}
\end{table*}

\begin{table*}[h]
\centering
\begin{tabular}{r|cccccc}
\hline
\textbf{Model}               &\textbf{View}  & \textbf{Accuracy}         & \textbf{Precision}        & \textbf{Recall}           & \textbf{F1-score} \\ \hline
Baseline (ResNet-50)         &Surface & 81.20 $\pm$ 6.02          & 83.80 $\pm$ 5.24          & 81.13 $\pm$ 5.85          & 81.16 $\pm$ 6.20  \\
Siamese ($k$=3, size=128)      &Surface & 80.21 $\pm$ 2.85          & 83.40 $\pm$ 2.85          & 80.21 $\pm$ 2.85          & 79.97 $\pm$ 2.86   \\
Triplet ($k$=7, size=512)      &Surface & 79.64 $\pm$ 4.22          & 81.73 $\pm$ 3.92          & 79.65 $\pm$ 4.22          & 79.48 $\pm$ 4.29  \\
GEMINI ($k$=1, size=512)       &Surface & 83.91 $\pm$ 1.93          & 86.43 $\pm$ 0.95          & 83.91 $\pm$ 1.93          & 84.15 $\pm$ 1.79  \\
GEMINI ($k$=1000, size=128)    &Surface & \textbf{89.16 $\pm$ 0.83} & \textbf{89.81 $\pm$ 0.89} & \textbf{89.16 $\pm$ 0.83} & \textbf{88.99 $\pm$ 0.86} \\ \hline

Baseline (ResNet-50)          &Section & \textbf{88.80 $\pm$ 6.02}     & \textbf{89.90 $\pm$ 5.24}               & \textbf{88.30 $\pm$ 5.85}          & \textbf{88.20 $\pm$ 6.20}            \\
Siamese ($k$=5, size=128)       &Section & 82.39 $\pm$ 2.73   & 84.17 $\pm$ 2.33              & 82.39 $\pm$ 2.73           & 82.29 $\pm$ 3.01           \\
Triplet ($k$=10, size=64)       &Section & 80.64 $\pm$ 1.34   & 83.59 $\pm$ 1.49              & 80.64 $\pm$ 1.34           & 80.63 $\pm$ 1.35           \\
GEMINI ($k$=1, size=16)         &Section & 79.08 $\pm$ 4.67    & 82.70 $\pm$ 3.51              & 79.08 $\pm$ 4.67          & 79.16 $\pm$ 4.90                     \\
GEMINI ($k$=1000, size=16)      &Section & 82.03 $\pm$ 4.77    & 83.07 $\pm$ 4.07              & 82.03 $\pm$ 4.77          & 82.20 $\pm$ 4.56                    \\ \hline

\end{tabular}
\caption{Performance comparison on surface (SUR) and section (SEC) dataset.}
\label{tab:comparison-models}
\end{table*}



\subsection{Evaluation of other DML approaches}
Besides the comparison with traditional DL classification models, the performance of the proposed GEMINI based approach was also compared to that of state-of-the-art DML methods using the kidney stone data. Two reference DML models were considered, namely the Siamese and Triplet networks, trained on both surface and section patches.

The experimental setup of both models remained the same with respect to the baseline and proposed models: no data augmentation or additional pre-processing, test with different embedding sizes, and averaged results over multiple iterations with random initialized seeds.

The implementation of both models was based on semi-hard sample mining to ensure that, during training, the most informative samples are used to obtain the best possible performance. Both the sample mining (using default margin values) and loss functions (Contrastive and Triplet margin loss) were implemented using the repository by \cite{musgrave2020pytorch}. Nevertheless, as mentioned in Section \ref{DeepMetricLearning}, the use of pair or triplet samples is resource intensive. Thus, with the chosen models, it was necessary to use more than one GPU (unlike the baseline model and proposal of this contribution). The embedding size and batch size parameter values are limited by the available GPU memory. This is particularly the case for the batch size which is the most resource demanding.

\section{Results on endoscopic data}
\label{experiments}

The average performance was assessed across 10 experiments for each DL-network configuration (i.e., for a given number of $k$-NN, embedding size, etc.) with a 95\% confidence interval. 

Table \ref{tab:comparison-stateoftheart}, which provides the network configurations achieving the highest average performance evaluated on the surface (SUR) and section (SEC) datasets, was established using  Fig.~\ref{fig:knn} giving recall values for numerous combinations of embedding-size and $k$-NN pairs.
A general trend can be observed in Table \ref{tab:comparison-stateoftheart}: the performance (precision, recall and F1-score values) increases with larger $k$ numbers of nearest neighbors
(bold numbers in Table \ref{tab:comparison-stateoftheart} and the next tables highlight the best criterion values).  It can also be noticed that in the best-performing configurations, the embedding size in the proposed model is equal to or smaller than that of the DML-models like the Siamese and Triplet networks.

\subsection{Evaluation of surface images}

As seen in Table \ref{tab:comparison-models}, the GEMINI approach globally exhibits the best performance on the SUR dataset.  The best GEMINI configuration (i.e., that with $k = 1000$ and an embedding-size of 128) led to an accuracy of 89.16\%, this quality criterion value being increased by 9.8\% in comparison to that of the base line model (ResNet-50, accuracy of 81.2\%) and by 11,95\% with respect to that of the Triplet Network (accuracy of 79.64\%).
It is also noticeable in Table \ref{tab:comparison-models}, that the best accuracy value of the GEMINI model was obtained for $k$ = 1000 and an  embedding size of 128, while with a decreasing number $k$ of nearest neighbors the accuracy diminishes (the accuracy falls to 83.91\% for $k$ =1 and with an embedding size of 512)    

For the SUR dataset (see Figs~\ref{fig:DMLSUR}.(a-3) and \ref{fig:DMLSUR}.(b-3)), the proposed GEMINI model achieves a high class compactness and a rather clear separation between the point clusters of the classes. It is particularly noticeable that the whewellite (WW) class corresponds to a compact cluster both in train and test spaces, while the uric acid (UA) and struvite (STR) classes tend to be partly overlapped. The separability of the whewellite (WW) class is highlighted by its cluster distance with respect to that of the other classes and its compactness corresponding to a small intra-class variability.  Some classes are partly overlapped. For the struvite (STR) and brushite (BRU) classes, this overlap can be explained by their visual similarities in terms of textures and colors. 

Rather stronger overlaps can be observed in the Triplet network (see Figs~\ref{fig:DMLSUR}.(a-2) and \ref{fig:DMLSUR}.(b-2)) since three classes (STR, BRU and WW) share common regions in the reduced embedding spaces (in the GEMINI model, classes are partly and pairwise overlap). 

For the Siamese network (see Figs.~\ref{fig:DMLSUR}.(a-1) and \ref{fig:DMLSUR}.(b-1)), it can be noticed that the whewellite (WW) and cystine (CYS) classes can be visually separated from the other four classes, but their compactness is weak. Even if the AU, STR, BRU and weddelite classes are more compact, they overlapped or are at least close to each other in the reduced embedding spaces.  

\begin{figure*}[] 
    \centering
    \includegraphics[width=0.90\linewidth]{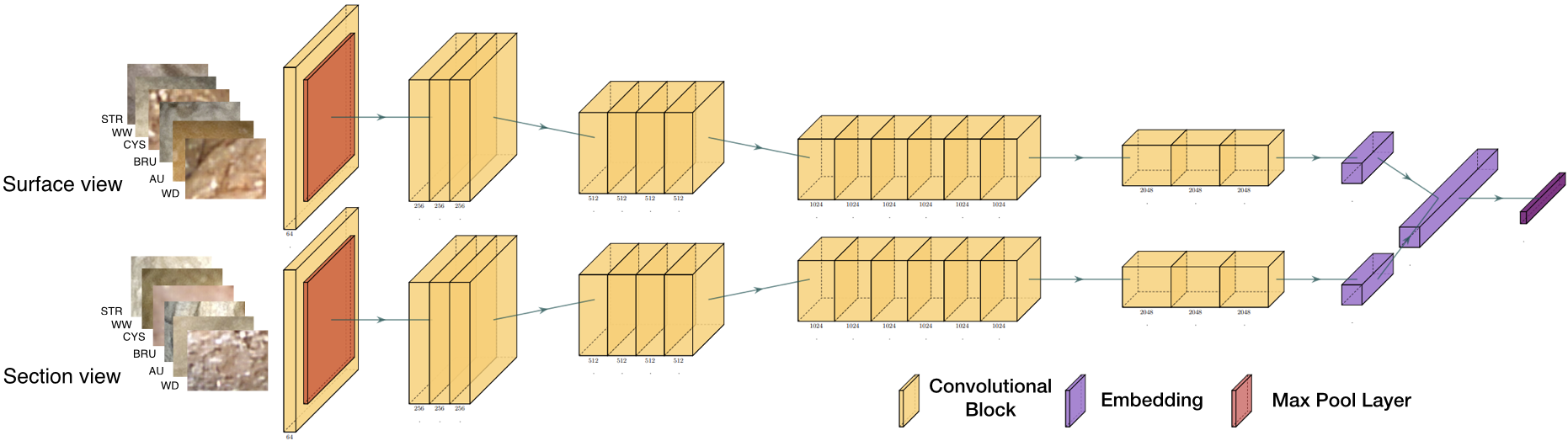}
    \hspace{0.5cm}
    \vspace{-0.1cm}
    \caption{Proposed multi-view model. The first part of the model corresponds to the duplicated feature extraction layers from the metric learning model. These layers are followed by the fusion layer, which combines information from the two views (i.e., from the two image types). The fused feature map is then connected to the classification layer.}
    \label{fig:fusion_model}
\end{figure*}
\subsection{Evaluation on section images}
For the SEC dataset (see Table \ref{tab:comparison-models}), the baseline model exhibits the best performance for all four quality criteria. 
Indeed,the 88.8\% accuracy of ResNet-50 model is 7.78\% and 12.29\% higher than that of the Siamese and GEMINI (with $k$ = 1 and an embedding size of 16), respectively.  

The proposed GEMINI model with $k$ = 1000 and an embedding-size of 16 has an accuracy of 82.03\% which is 1.72\% higher than that of the Triplet model (accuracy of 80.64\%) and  3.7\% higher than the GEMINI model with configuration $k=1$ and embedding-size of 16 (accuracy of 79.08\%).

As visible in Figs.~\ref{fig:DMLSEC}.(c-1) and \ref{fig:DMLSEC}.(c-2) for the SEC dataset, there is a noticeable difference in the cluster (shape, size and position) between the training and test embedding spaces generated by the proposed GEMINI model. The same observation is valid for the spaces generated by the Siamese (see Figs.~\ref{fig:DMLSEC}.(a-1) and \ref{fig:DMLSEC}.(b-1) and Triplet (see Figs.~\ref{fig:DMLSEC}.(a-2) and \ref{fig:DMLSEC}.(b-2)) networks. This systematic difference between the class clusters in the train and test spaces produced by each network indicates that the training and test data did not carry a similar information distribution. 

By visually comparing the training and test embedding spaces in Figs.~\ref{fig:DMLSUR} and \ref{fig:DMLSEC}, the surface view information can lead to more efficient class separation than the section view of the kidney stone data, whatever the model (GEMINI, Triplet, or Siamese network). This hypothesis is numerically confirmed in Table \label{tab:comparison-stateoftheart} for the proposed method for all values of $k$.

\begin{table*}[]
\begin{tabular}{@{}rccccl@{}}
\toprule
\textbf{Method}         & \textbf{Accuracy}    & \textbf{Precision}   & \textbf{Recall}      & \textbf{F1-Score}    & \textbf{Fusion strategy}                      \\ \midrule
\cite{martinez2020}        & 52.7 $\pm$ 18.9          & 55.2 $\pm$ 30.1          & 52.7 $\pm$ 20.7          & 52.9 $\pm$ 21.3          & Concatenation             \\
\cite{black2020}           & 80.1 $\pm$ 13.8          & 81.2 $\pm$ 15.4          & 80.1 $\pm$ 15.1          & 80.1 $\pm$ 13.7          & Mixed views                        \\
\cite{estrade2022towards}         & 70.1 $\pm$ 22.3          & 72.0 $\pm$ 23.0          & 70.1 $\pm$ 24.5          & 69.9 $\pm$ 23.3          & Mixed views                        \\
\cite{lopez2021assessing}           & 81.2 $\pm$ 06.0          & 83.8 $\pm$ 05.2          & 81.1 $\pm$ 05.8          & 81.1 $\pm$ 06.2          & Mixed views                        \\
\cite{lopez2022boosting}          & 85.6 $\pm$ 0.10          & 86.8 $\pm$ 0.20          & 85.6 $\pm$ 00.1          & 85.4 $\pm$ 00.1          & Mixed views                        \\ \midrule
\textbf{Proposed strategy 1} & 86.0 $\pm$ 02.2 & 87.3 $\pm$ 01.7 & 86.0 $\pm$ 02.2 & 85.3 $\pm$ 03.4 & Concatenation             \\
\textbf{Proposed strategy 2} & \textbf{88.7 $\pm$ 02.0} & \textbf{90.0 $\pm$ 01.1} & \textbf{88.7 $\pm$ 02.0} & \textbf{88.5 $\pm$ 02.3} & Stack \& Max-pool \\ \bottomrule
\end{tabular}
\caption{State-of-the-art comparison for the classification methods using both views simultaneously. In this work, two different methods were used to perform the fusion of both views (strategy 1 with concatenation and strategy 2 with stack and max-pool layers).}
\label{tab:results_sota}
\end{table*}

\begin{figure*}[h!] 
    \centering
    \subfloat[Train]{
    \label{fig:FusionSectionTrainSpace}\includegraphics[width=0.40\linewidth]{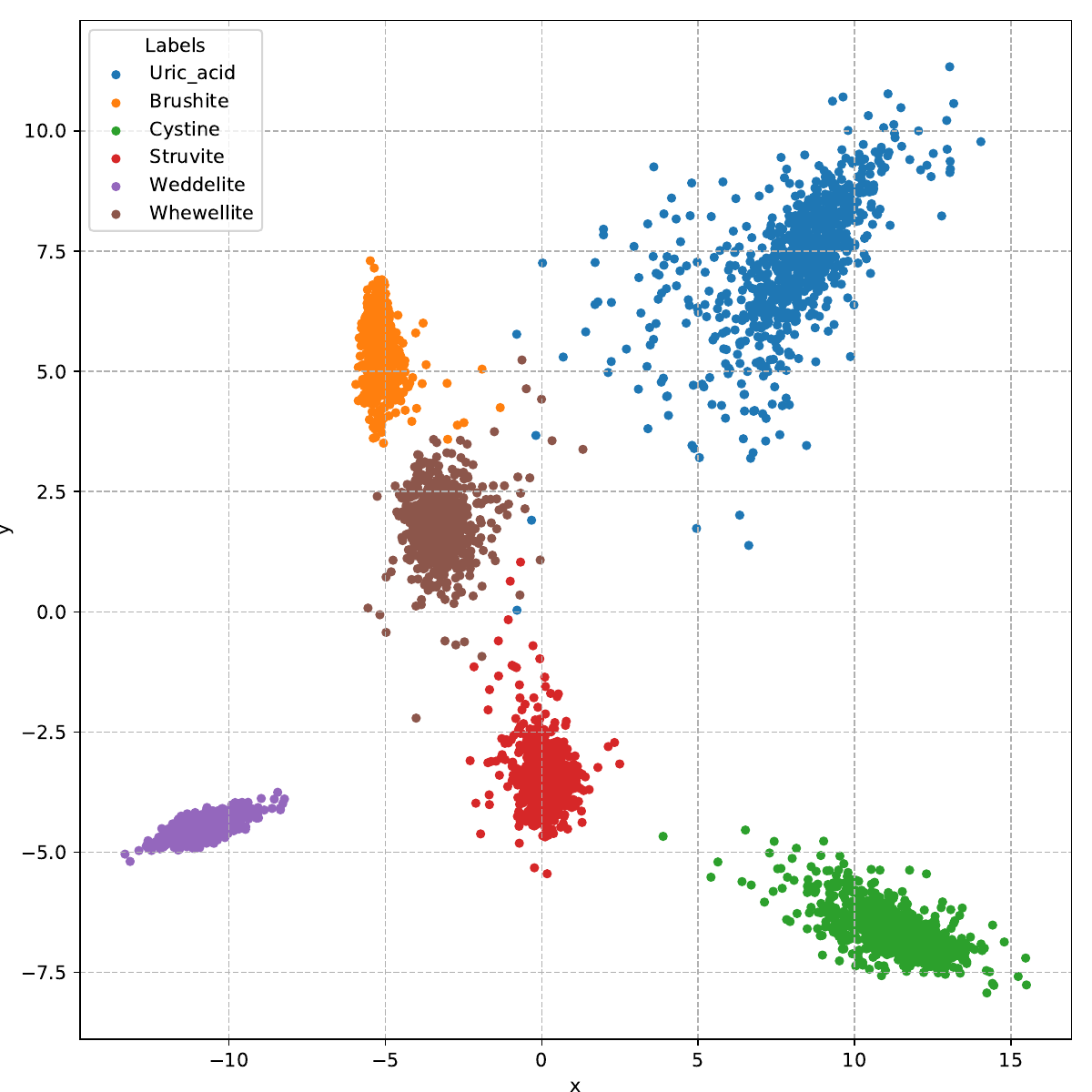}}
    \hspace{0.5cm}
    \subfloat[Test]{
    \label{fig:FusionSectionTestSpace}\includegraphics[width=0.40\linewidth]{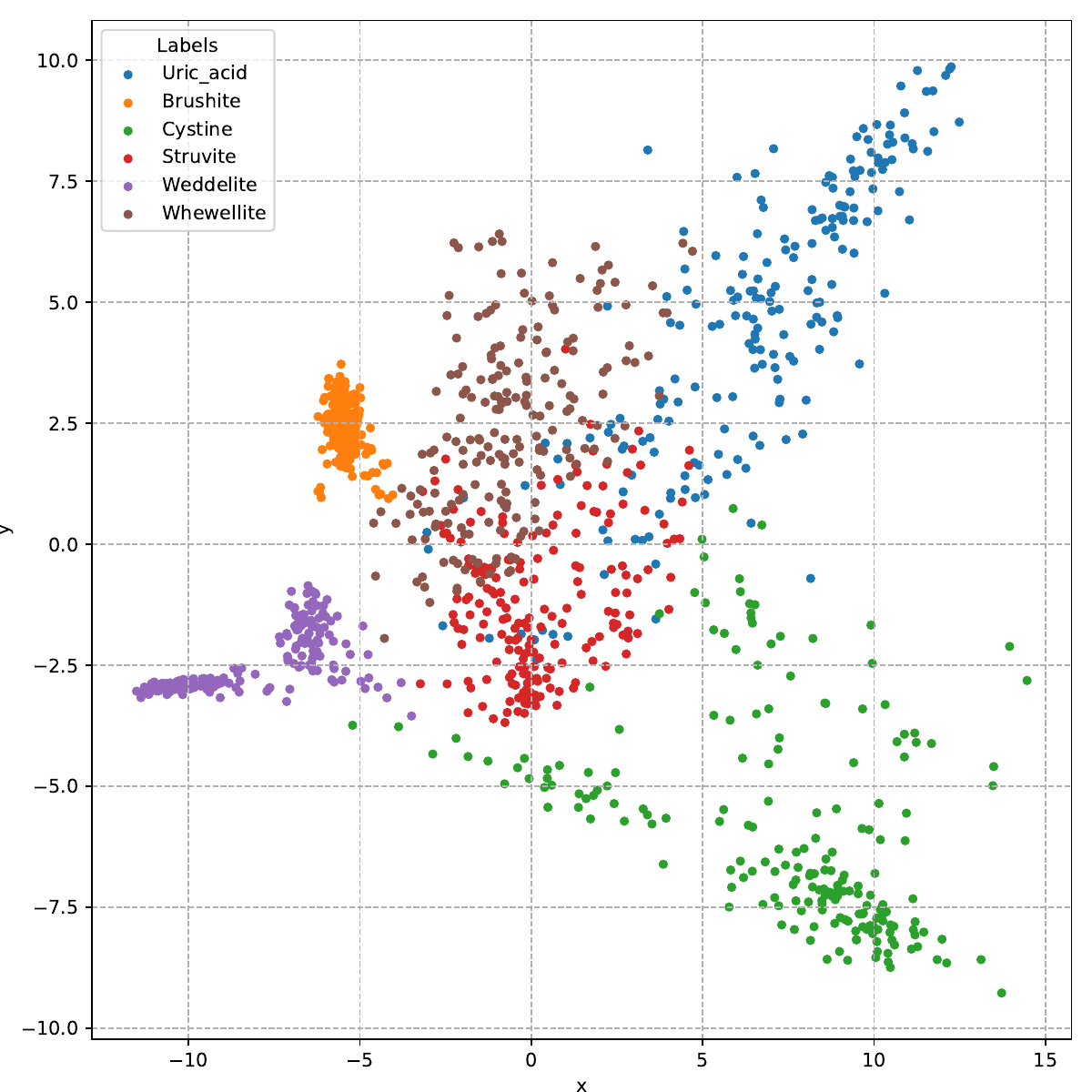}}
    \vspace{-0.1cm}
    \caption{Fusion method embedding space in the mixed dataset. The \textbf{(a)} train and \textbf{(b)} test embedding spaces were extracted from the penultimate layer output as feature descriptors and then reduced to 2-D representations via PCA.}
    \label{fig:FusionSectionSpace}
\end{figure*}

\subsection{Fusion Strategies \label{fusion-estrategies}}
 
The best GEMINI based models separately obtained for the SUR and SEC datasets can be more optimally exploited by combining them through fusion strategies. The two models used in this experiment were those with the embedding size values that generated the highest accuracy in SUR and SEC views, i.e., those with embedding size 128 (SUR, in Table 3) and 16 (SEC in Table 3), respectively. 
Two strategies were tested to fuse the information of the two views. In the first strategy, the features from each network are concatenated and then connected to a classifier. The resulting embedding size corresponds to the sum of both models, i.e., embedding-size of 144 = 128 (SUR) + 16 (SEC). In the second strategy, the features generated by each network are channel-wise stacked together to create an embedding of shape $(H, W, V)$ where $V$ refers to as  the number of modalities that are being stacked. A max-pooling operation is performed to reduce the number of channels. Since the shape of the embeddings from each model is different, the values from the previous layer are considered instead. The embedding size generated from this approach is $(H, W, 1)$, which is finally connected to a classifier.

Each configuration of the fusion model was trained for 10 epochs with a learning rate of $1e-3$. The proposed fusion model is sketched in Fig.~\ref{fig:fusion_model}
 
The results obtained in this contribution are compared to those of the state-of-the-art in Table \ref{tab:results_sota}. It can be observed in this table that the use of fusion layers enhances the overall kidney stone identification performance for both strategies.
The stack and max-pool layer strategy led to the best performance for all four quality criteria. The accuracy criterion value was notably increased by 3.6\% by strategy 2 in comparison to the most performing state-of-the-art contribution (\cite{lopez2022boosting}). Similarly, the concatenation approach reached a performance leading to an accuracy increase of 63.3\% in comparison to the method in (\cite{martinez2020}) that extracts handcrafted features.

In the train embedding space  of the best fusion approach (strategy 2 in Table \ref{tab:results_sota}), the six classes correspond to clusters 
(see Fig.~\ref{fig:FusionSectionSpace}.(a)) which are compact and almost without overlap.  Even though there is a shift in the distribution between the training and test embedding spaces which is particularly noticeable in the weddelite (WD) and cystine (CYS) classes in Fig.~\ref{fig:FusionSectionSpace}.(b), the visual separability in both 2D reduced spaces explain why strategy 2 in Table \ref{tab:results_sota} achieves an high performance in the kidney stone type recognition.

\section{Discussion} 
\label{discussion}
This section discusses the performance of the proposed GEMINI scheme. In comparison to the state-of-the-art methods, the proposed approach demonstrates systematically a high performance for all three datasets and for all four quality criteria given in Table \ref{tab:results_sota}. Concerning the SUR dataset, both variants of the fusion strategies outperformed the performance of the other DML and traditional DL classification methods. This result is promising since the ex-vivo kidney-stones were acquired with endoscopes actually used in ureteroscopy and acquisition conditions were realistic (the quality and available information given by the ex-vivo data are very close to that of in-vivo data).

It can also be noticed that, in the SEC dataset, the proposed model has comparable to moderately superior results than other DML models. While the best average performance (over 6 classes) is found in the tested traditional DL classification method, the DML models show results with more narrow confidence intervals (Table \ref{tab:comparison-models}) . Moreover, there is room for improvement by testing with different data splits that balance test and training data.

\begin{figure*}
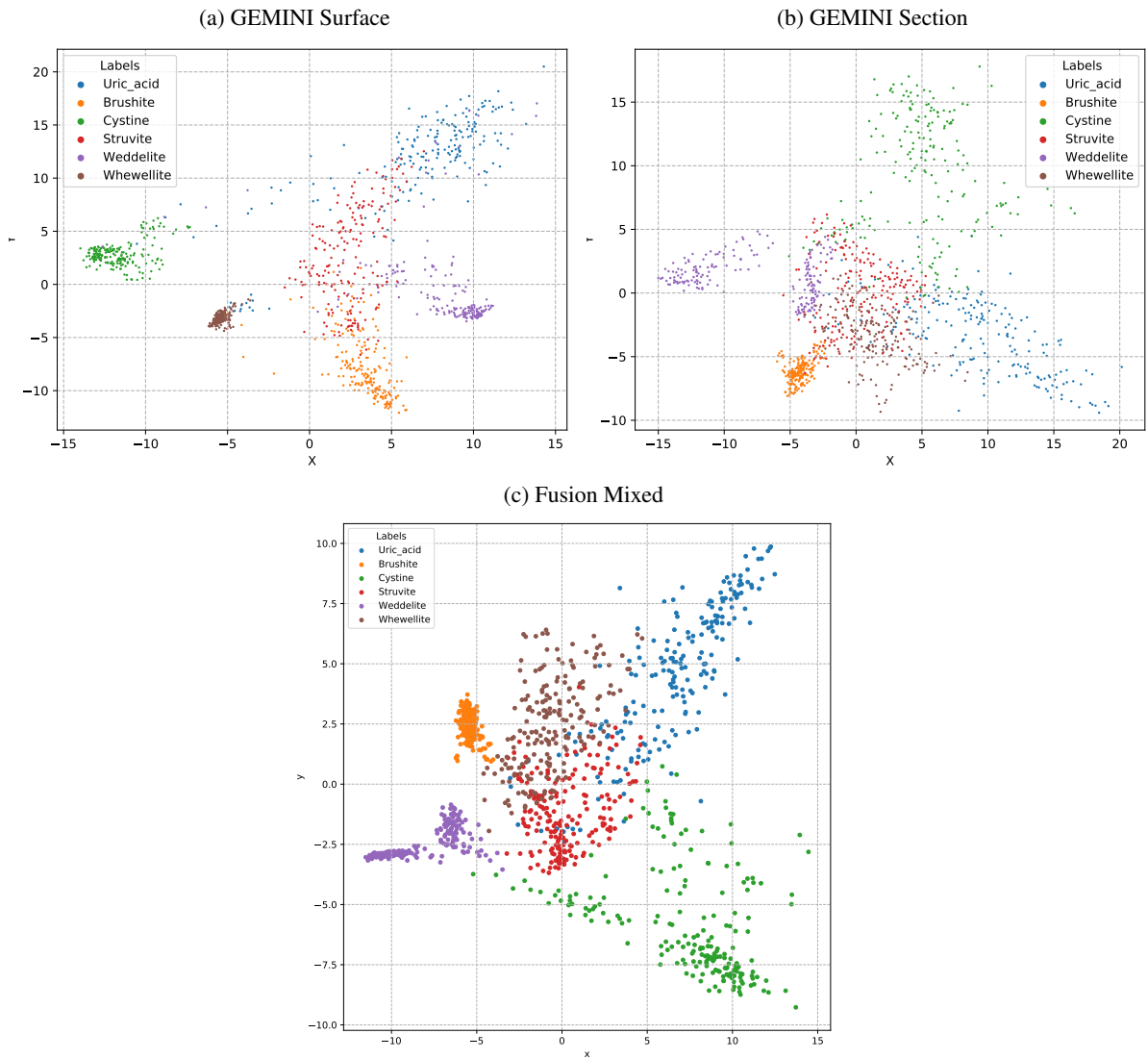

    \centering
    
    \begin{subfigure}[b]{0.45\textwidth}
    \label{fig:GEMINISurfaceEmbeddingSpace}
    \caption{GEMINI Surface}
    \includegraphics[width=1.0\linewidth]{images/SurfaceTestSpace.pdf}
    \end{subfigure}
    \begin{subfigure}[b]{0.45\textwidth}
    \label{fig:GEMINISectionEmbeddingSpace}
    \caption{GEMINI Section}
    \includegraphics[width=1.0\linewidth]{images/SectionTestSpace.pdf}
    
    \end{subfigure}
    \begin{subfigure}[b]{0.45\textwidth}
    \label{fig:FusionMixedEmbeddingSpace}
    \caption{Fusion Mixed}
    \includegraphics[width=0.95\linewidth]{images/MixedTestSpace.pdf}
    \end{subfigure}

    \caption{Comparison between \textbf{a)} GEMINI surface, \textbf{b)} GEMINI section, and \textbf{c)} Fusion mixed embedding space.}
    \label{fig:ComparisonSurSecMix}
\end{figure*}

In the fusion approach, fusion layers were used to combine the features extracted for the two kidney stones views (SUR and SEC), with the aim to generate more discriminative information (see Fig.~\ref{fig:ComparisonSurSecMix}). The two fusion approaches demonstrated an improvement in performance, along with small confidence intervals.  Nevertheless, the proposed  fusion approach could be enhanced by completing the networks, for instance with attention mechanisms that potentially can improve the classification accuracy, as shown in \cite{villalvazo2022improved}.

\section{Conclusion \label{conclusion}}
This contributions studied the appropriateness of Deep Metric Learning (DML) models for kidney stone classification and compared the performance of such networks to that of other kidney stone classification approaches. The proposed DML architecture uses local and global information contributing to a manifold generalization, and leading to more discriminant feature spaces in the case of surface kidney stones datasets . This contribution also shows that significant embeddings can be learned without a strict sample selection phase.

The results obtained demonstrate that DML models in kidney stone classification tasks can be a first step towards a reliable and automated identification tool in ureteroscopy. Given the ability to generate low-dimensional data representations that make them also useful for data visualization, DLM models  could also be used as decision support or exploration of sample features to assist urologists in the kidney stone identification. 
Similarly, DML models can be extended to other tasks in the medical area where there are a limited number of samples or environmental variables (backgrounds, poses, illuminations, etc.)  that limit or hinder a correct classification.

\section*{Acknowledgments}
The authors wish to acknowledge the Mexican Council for Science and Technology (CONACYT) for the support in terms of postgraduate scholarships in this project, and the Data Science Hub at Tecnologico de Monterrey for their support on this project. 
This work has been supported by Azure Sponsorship credits granted by Microsoft's AI for Good Research Lab through the AI for Health program.

\section*{Compliance with ethical approval}
The images were captured in medical procedures following the ethical principles outlined in the Helsinki Declaration of 1975, as revised in 2000, with the consent of the patients.

\bibliographystyle{cas-model2-names}

\bibliography{cas-refs}

\end{document}